\documentclass[11pt]{article}
\input{glyphtounicode}\pdfgentounicode=1 %

\usepackage[top=0.85in,bottom=1in,left=1in,right=1in]{geometry}

\usepackage{amsmath}
\usepackage{amssymb}
\usepackage{array}
\usepackage{tabularx}
\usepackage{booktabs}
\usepackage{graphicx}
\usepackage{float}
\usepackage[section]{placeins}
\usepackage{hyperref}
\usepackage{xurl}
\usepackage{titling}
\usepackage{textcomp}
\usepackage{adjustbox}
\usepackage{subcaption}

\usepackage[style=apa, backend=biber]{biblatex}
\newcommand{\AccBetween}{84.5}
\newcommand{\AccFar}{53.0}
\newcommand{\AccFarHi}{55.0}
\newcommand{\AccFarLo}{51.0}
\newcommand{\AccMid}{65.4}
\newcommand{\AccMidHi}{67.2}
\newcommand{\AccMidLo}{63.7}
\newcommand{\AccNative}{94.3}
\newcommand{\AccNativeHi}{96.1}
\newcommand{\AccNativeLo}{92.2}
\newcommand{\AccNear}{84.7}
\newcommand{\AccNearHi}{86.1}
\newcommand{\AccNearLo}{83.2}
\newcommand{\AccWithin}{91.5}
\newcommand{\CovGlasgowPlan}{89}
\newcommand{\CovThingsNPlan}{64}
\newcommand{\CovThingsPairs}{3{,}400}
\newcommand{\CovThingsPairsPlan}{3{,}486}
\newcommand{\CvclFar}{42.8}
\newcommand{\CvclMid}{51.1}
\newcommand{\CvclNative}{67.5}
\newcommand{\CvclNear}{64.7}
\newcommand{\CvclTierRho}{\ensuremath{-}.84}
\newcommand{\DisBeta}{\ensuremath{-}0.212}
\newcommand{\DisPRperc}{0.245}
\newcommand{\DisPRsem}{0.002}
\newcommand{\DisPperc}{1.87 \times 10^{-7}}
\newcommand{\DisPsem}{0.660}
\newcommand{\DisSuperB}{0.015}
\newcommand{\DisSuperP}{0.528}
\newcommand{\DistBasSup}{.861}
\newcommand{\DistOneSup}{.880}
\newcommand{\DistSubSup}{.892}
\newcommand{\DistSupMax}{.892}
\newcommand{\DistSupMin}{.845}
\newcommand{\DistSupSup}{.845}
\newcommand{\DmemFar}{.694}
\newcommand{\DmemMid}{.559}
\newcommand{\DmemNear}{.359}
\newcommand{\FirstSessDiff}{one}
\newcommand{\HFiveHundredN}{85}
\newcommand{\HFiveHundredP}{.32}
\newcommand{\HFiveHundredRho}{\ensuremath{-}.11}
\newcommand{\HFiveTenPBH}{.0026}
\newcommand{\HFiveTenRho}{\ensuremath{-}.89}
\newcommand{\HFourB}{\ensuremath{-}3.226}
\newcommand{\HFourBshort}{\ensuremath{-}3.23}
\newcommand{\HFourExclB}{\ensuremath{-}3.12}
\newcommand{\HFourExclHi}{\ensuremath{-}3.05}
\newcommand{\HFourExclLo}{\ensuremath{-}3.19}
\newcommand{\HFourExclPersons}{464}
\newcommand{\HFourFirstB}{\ensuremath{-}3.17}
\newcommand{\HFourFirstHi}{\ensuremath{-}3.10}
\newcommand{\HFourFirstLo}{\ensuremath{-}3.23}

\newcommand{\HFourHi}{\ensuremath{-}3.164}
\newcommand{\HFourLo}{\ensuremath{-}3.288}
\newcommand{\HFourNobs}{172{,}800}
\newcommand{\HFourPersons}{520}
\newcommand{\HFourPreregB}{\ensuremath{-}3.17}
\newcommand{\HFourPreregHi}{\ensuremath{-}3.10}
\newcommand{\HFourPreregLo}{\ensuremath{-}3.24}
\newcommand{\HFourPreregPersons}{444}
\newcommand{\HFourRepeaters}{56}
\newcommand{\HFourSessions}{600}
\newcommand{\HFourTargetB}{\ensuremath{-}3.16}
\newcommand{\HFourTargetExclB}{\ensuremath{-}3.11}
\newcommand{\HFourTargetExclHi}{\ensuremath{-}2.96}
\newcommand{\HFourTargetExclLo}{\ensuremath{-}3.26}
\newcommand{\HFourTargetHi}{\ensuremath{-}3.02}
\newcommand{\HFourTargetLo}{\ensuremath{-}3.30}
\newcommand{\HFourTrials}{7{,}200}
\newcommand{\HOneCtenHi}{.897}
\newcommand{\HOneCtenLo}{.606}
\newcommand{\HOneCtenN}{45}
\newcommand{\HOneCtenRho}{.795}
\newcommand{\HOneHolmP}{.0005}
\newcommand{\HOneHsjHi}{.543}
\newcommand{\HOneHsjLo}{.456}
\newcommand{\HOneHsjN}{1{,}390}
\newcommand{\HOneHsjRho}{.501}
\newcommand{\HOneThingsHi}{.549}
\newcommand{\HOneThingsLo}{.496}
\newcommand{\HOneThingsN}{3{,}400}
\newcommand{\HOneThingsRho}{.523}
\newcommand{\HSixB}{.0047}
\newcommand{\HSixBHTests}{9}
\newcommand{\HSixBaseCvcl}{.618}
\newcommand{\HSixBaseDino}{.850}
\newcommand{\HSixHi}{.028}
\newcommand{\HSixHsjHundred}{.43}
\newcommand{\HSixLo}{\ensuremath{-}.018}
\newcommand{\HSixP}{.688}
\newcommand{\HSixRhoHundred}{.62}
\newcommand{\HSixRhoTen}{.75}
\newcommand{\HSixThingsHundred}{.68}
\newcommand{\HTwoHsjBeta}{\ensuremath{-}.034}
\newcommand{\HTwoHsjN}{32}
\newcommand{\HTwoHsjP}{.039}
\newcommand{\HTwoHsjPHolm}{.079}
\newcommand{\HTwoHsjR}{\ensuremath{-}.41}
\newcommand{\HTwoHsjRHi}{\ensuremath{-}.07}
\newcommand{\HTwoHsjRLo}{\ensuremath{-}.67}
\newcommand{\HTwoThingsBeta}{\ensuremath{-}.008}
\newcommand{\HTwoThingsN}{61}
\newcommand{\HTwoThingsP}{.374}

\newcommand{\HTwoThingsR}{\ensuremath{-}.10}
\newcommand{\HTwoThingsRHi}{.16}
\newcommand{\HTwoThingsRLo}{\ensuremath{-}.34}
\newcommand{\IncChi}{3{,}017.0}
\newcommand{\IncChiRE}{3{,}695.4}
\newcommand{\IncDAIC}{3{,}015.0}
\newcommand{\IntBasB}{\ensuremath{-}0.86}
\newcommand{\IntBasHi}{\ensuremath{-}0.79}
\newcommand{\IntBasLo}{\ensuremath{-}0.94}
\newcommand{\IntChi}{2{,}134.2}
\newcommand{\IntRefB}{\ensuremath{-}2.92}
\newcommand{\IntRefHi}{\ensuremath{-}2.85}
\newcommand{\IntRefLo}{\ensuremath{-}3.00}
\newcommand{\IntSubB}{0.10}
\newcommand{\IntSubHi}{0.16}
\newcommand{\IntSubLo}{0.04}
\newcommand{\IntSupB}{\ensuremath{-}1.29}
\newcommand{\IntSupHi}{\ensuremath{-}1.22}
\newcommand{\IntSupLo}{\ensuremath{-}1.36}
\newcommand{\LvlBasB}{\ensuremath{-}3.21}
\newcommand{\LvlBasHi}{\ensuremath{-}3.03}
\newcommand{\LvlBasLo}{\ensuremath{-}3.40}
\newcommand{\LvlSubB}{\ensuremath{-}4.22}
\newcommand{\LvlSubHi}{\ensuremath{-}3.98}
\newcommand{\LvlSubLo}{\ensuremath{-}4.46}
\newcommand{\LvlSupB}{\ensuremath{-}0.94}
\newcommand{\LvlSupHi}{\ensuremath{-}0.62}
\newcommand{\LvlSupLo}{\ensuremath{-}1.25}
\newcommand{\MantelPerms}{10{,}000}
\newcommand{\MantelPmin}{.0001}
\newcommand{\MapHsjClasses}{68}
\newcommand{\MapThingsClasses}{83}
\newcommand{\NPairsFar}{4}
\newcommand{\NPairsMid}{3}
\newcommand{\NPairsNear}{3}
\newcommand{\NWithin}{6}
\newcommand{\OrdIntChi}{29.12}
\newcommand{\OrdIntP}{< .001}
\newcommand{\OrdIntPHolm}{< .001}
\newcommand{\OrdNobs}{172{,}800}
\newcommand{\OrdPersons}{520}
\newcommand{\OrdZdB}{0.20}
\newcommand{\OrdZdChi}{24.59}
\newcommand{\OrdZdHi}{0.28}
\newcommand{\OrdZdLo}{0.12}
\newcommand{\OrdZdP}{< .001}
\newcommand{\OrdZdPHolm}{< .001}

\newcommand{\OrdZdTriB}{0.06}
\newcommand{\OrdZdTriP}{= .25}
\newcommand{\PlanHash}{9b50995eb5e8}
\newcommand{\PlanHashAnon}{b1a70c010fa5}
\newcommand{\SDdAll}{.219}
\newcommand{\SDdTarget}{.296}
\newcommand{\SelBasSup}{1.8}
\newcommand{\SelOneSup}{2.6}
\newcommand{\SelSubSup}{1.1}
\newcommand{\SelSupOverall}{20.7}
\newcommand{\SelSupSup}{77.5}
\newcommand{\SupCBasB}{\ensuremath{-}0.49}
\newcommand{\SupCBasHi}{\ensuremath{-}0.22}
\newcommand{\SupCBasLo}{\ensuremath{-}0.76}
\newcommand{\SupCSubB}{\ensuremath{-}1.05}
\newcommand{\SupCSubHi}{\ensuremath{-}0.73}
\newcommand{\SupCSubLo}{\ensuremath{-}1.37}
\newcommand{\SupCSupB}{7.11}
\newcommand{\SupCSupHi}{7.36}
\newcommand{\SupCSupLo}{6.87}
\newcommand{\SupCondChi}{18{,}907.2}
\newcommand{\SupCondP}{< .001}
\newcommand{\SupDistChi}{15.14}
\newcommand{\SupDistP}{< .001}
\newcommand{\SupImages}{12}
\newcommand{\SupNobs}{28{,}800}
\newcommand{\SupPersons}{520}

\newcommand{\SupSlZdB}{\ensuremath{-}0.28}
\newcommand{\SupSlZdHi}{\ensuremath{-}0.13}
\newcommand{\SupSlZdLo}{\ensuremath{-}0.42}
\newcommand{\SupZdB}{\ensuremath{-}0.28}
\newcommand{\SupZdHi}{\ensuremath{-}0.13}
\newcommand{\SupZdLo}{\ensuremath{-}0.42}
\newcommand{\WnCorrBeta}{\ensuremath{-}0.223}
\newcommand{\WnCorrNQ}{35}
\newcommand{\WnCorrPsem}{0.718}
\newcommand{\WnCorrQFour}{.863}
\newcommand{\WnCorrQThree}{.838}
\newcommand{\WnFrozenBeta}{\ensuremath{-}0.212}
\newcommand{\WnFrozenPsem}{0.660}
\newcommand{\WnNCorrected}{21}
\newcommand{\WnNTouch}{42}
\newcommand{\XRho}{\ensuremath{-}.72}
\newcommand{\XSlope}{\ensuremath{-}6.93}
\newcommand{\XSlopeHi}{\ensuremath{-}6.56}
\newcommand{\XSlopeLo}{\ensuremath{-}7.29}
\newcommand{\XUnits}{96}
\newcommand{\HFourLoshort}{\ensuremath{-}3.29}
\newcommand{\HFourHishort}{\ensuremath{-}3.16}

\newcommand{\VtTurtleSessions}{600}
\newcommand{\VtTurtleAnimalsK}{431}
\newcommand{\VtTurtleVehiclesK}{1}
\newcommand{\VtTurtleVegK}{3}
\newcommand{\VtTurtleOnlyAnimals}{431}
\newcommand{\VtTurtleOnlyVehicles}{1}
\newcommand{\VtTurtleBoth}{0}
\newcommand{\VtTurtleNeither}{168}
\newcommand{\VtPersons}{520}
\newcommand{\VtXSameBasicPct}{1.83}
\newcommand{\VtXInDenom}{7{,}200}
\newcommand{\VtXBroadPct}{77.49}
\newcommand{\VtXOutBroadPct}{0.34}
\newcommand{\VtXOutDenom}{28{,}800}
\newcommand{\VtXViTIn}{10}

\newcommand{\VtXInN}{12}
\newcommand{\VtXViTOut}{0}
\newcommand{\VtXDinoBOut}{20}
\newcommand{\VtXOutN}{48}
\newcommand{\VtXPerCell}{600}
\newcommand{\VtXJumpMin}{8.33}
\newcommand{\VtXJumpMax}{66.67}
\newcommand{\VtXJumpAdult}{76.43}
\newcommand{\VtXEthreeAnimals}{4}
\newcommand{\VtXEthreeVeg}{4}
\newcommand{\VtXEthreeVeh}{0}
\newcommand{\VtXRailcarK}{374}
\newcommand{\VtPcRandB}{\ensuremath{-}1.10}
\newcommand{\VtPcRandLo}{\ensuremath{-}1.22}
\newcommand{\VtPcRandHi}{\ensuremath{-}0.99}
\newcommand{\VtPcPixB}{0.12}
\newcommand{\VtPcPixLo}{0.04}
\newcommand{\VtPcPixHi}{0.19}
\newcommand{\VtPcViTB}{\ensuremath{-}4.30}
\newcommand{\VtPcViTLo}{\ensuremath{-}4.46}
\newcommand{\VtPcViTHi}{\ensuremath{-}4.13}
\newcommand{\VtPcRefB}{\ensuremath{-}3.16}
\newcommand{\VtPcRefLo}{\ensuremath{-}3.30}
\newcommand{\VtPcRefHi}{\ensuremath{-}3.02}
\newcommand{\VtPcMaeB}{\ensuremath{-}0.20}
\newcommand{\VtPcMaeLo}{\ensuremath{-}0.30}
\newcommand{\VtPcMaeHi}{\ensuremath{-}0.10}
\newcommand{\VtPcNobs}{57{,}600}
\newcommand{\VbPcCvclB}{\ensuremath{-}0.62}
\newcommand{\VbPcCvclLo}{\ensuremath{-}0.71}
\newcommand{\VbPcCvclHi}{\ensuremath{-}0.52}

\newcommand{\VtPcTrainedNeg}{8}
\newcommand{\VtHfourNobs}{172{,}800}
\newcommand{\VbHfourAllNeg}{10}
\newcommand{\VbHfourRandB}{\ensuremath{-}2.68}
\newcommand{\VbHfourCvclB}{\ensuremath{-}3.61}
\newcommand{\VbHfourMaeB}{\ensuremath{-}7.56}

\newcommand{\VtOrdDelta}{0.020}
\newcommand{\VtOrdLo}{\ensuremath{-}0.205}
\newcommand{\VtOrdHi}{0.274}
\newcommand{\VtGlobalCells}{9}
\newcommand{\VtGlobalLabel}{G1}
\newcommand{\VtNexp}{12}

\newcommand{\VtDSupportWord}{six}
\newcommand{\VtDrMin}{\ensuremath{-}.30}
\newcommand{\VtDrMax}{\ensuremath{-}.65}
\newcommand{\VtDCvclR}{.015}
\newcommand{\VtDCvclLo}{\ensuremath{-}.18}
\newcommand{\VtDCvclHi}{.21}
\newcommand{\VtDMaeR}{\ensuremath{-}.004}
\newcommand{\VtDMaeLo}{\ensuremath{-}.20}
\newcommand{\VtDMaeHi}{.19}
\newcommand{\VtDN}{100}
\newcommand{\VtDallN}{4{,}950}
\newcommand{\VtHoneMin}{.23}
\newcommand{\VtHoneMax}{.91}
\newcommand{\VtHoneRandThings}{.11}
\newcommand{\VtHonePixThings}{.15}
\newcommand{\VtHoneRandHsj}{.11}
\newcommand{\VtHonePixHsj}{.20}
\newcommand{\VtHoneRandCten}{.53}
\newcommand{\VtHonePixCten}{.29}
\newcommand{\VtHonePixCtenP}{.0505}
\newcommand{\VbHoneRandThingsP}{.038}
\newcommand{\VbHoneRandThingsLo}{.08}
\newcommand{\VbHoneRandThingsHi}{.14}
\newcommand{\VbHoneRandHsjP}{.013}
\newcommand{\VbHoneRandHsjLo}{.05}
\newcommand{\VbHoneRandHsjHi}{.16}
\newcommand{\VbHoneRandCtenP}{.006}
\newcommand{\VbHoneRandCtenLo}{.25}
\newcommand{\VbHoneRandCtenHi}{.71}
\newcommand{\VtHoneNThings}{3{,}400}
\newcommand{\VtHoneNHsj}{1{,}390}
\newcommand{\VtHoneNCten}{45}
\newcommand{\VtCtrlOneObs}{1.000000000139778\times 10^{-6}}
\newcommand{\VtCtrlOneLine}{125}
\newcommand{\VtCtrlTwoMax}{6.39\times 10^{-7}}
\newcommand{\VtCtrlTwoPairs}{4{,}950}
\newcommand{\VtPatchPub}{79ed554c\ldots{}12f0}
\newcommand{\VtPatchFrz}{21125d6a\ldots{}f87d}
\newcommand{\VtPatchFrzLine}{31}
\newcommand{\VtPatchHdrLines}{5}
\newcommand{\VtFreezeTime}{18:00:36}
\newcommand{\VtFreezeDate}{2026-11-22}
\newcommand{\VtXtDate}{2026-11-22}
\newcommand{\VtXtTime}{11:33}

\newcommand{\VcHoneTrMin}{.39}
\newcommand{\VcHoneTrMax}{.70}
\newcommand{\VcHoneTrLoMin}{.36}
\newcommand{\VcHoneMaeThings}{.29}
\newcommand{\VcHoneMaeHsj}{.23}
\newcommand{\VcXSameSubPct}{1.06}
\newcommand{\VcXStep}{8.33}
\newcommand{\VcXNextStep}{75.00}
\newcommand{\VcXJumpKcMin}{25.00}
\newcommand{\VcXJumpKcMax}{58.33}
\newcommand{\VcXJumpKnnMin}{8.33}
\newcommand{\VcXJumpKnnMax}{33.33}
\newcommand{\VcCropList}{DINOv2-small, DINOv2-base, supervised ResNet-50, and the random ResNet-50}
\newcommand{\VcFullList}{CVCL, SigLIP, the supervised ViT, and MAE}
\newcommand{\VcHtwoRandHsjP}{.032}
\newcommand{\VcHtwoRandHsjR}{\ensuremath{-}.35}
\newcommand{\VcHtwoRandHsjLo}{\ensuremath{-}.63}
\newcommand{\VcHtwoRandHsjHi}{.00}
\newcommand{\VcMdeThings}{.35}
\newcommand{\VcMdeHsj}{.48}
\newcommand{\VcMdeHalfThings}{.25}
\newcommand{\VcMdeHalfHsj}{.35}
\newcommand{\VcHtwoAllThingsN}{3{,}400}
\newcommand{\VcHtwoAllHsjN}{1{,}390}
\newcommand{\VcImgTrSameSignWord}{seven}

\newcommand{\VcImgTrExclWord}{five}
\newcommand{\VcImgTrInclList}{DINOv2-small, DINOv2-base, and MAE}
\newcommand{\VcImgRandB}{\ensuremath{-}1.85}
\newcommand{\VcImgRandLo}{\ensuremath{-}3.29}
\newcommand{\VcImgRandHi}{\ensuremath{-}0.40}

\newcommand{\VcImgSdMin}{1.6}
\newcommand{\VcImgSdMax}{12.9}
\newcommand{\VcSupFrozenDistChi}{15.14}

\newcommand{\VcDmeanBetterN}{6}
\newcommand{\VcDminBetterN}{2}
\newcommand{\VcSupCondChiMinAll}{5{,}042}
\newcommand{\VcImgRzB}{\ensuremath{-}2.79}
\newcommand{\VcImgRzLo}{\ensuremath{-}5.88}
\newcommand{\VcImgRzHi}{0.30}
\newcommand{\VcImgRzSd}{7.76}

\newcommand{\VcSupImgChi}{0.25}
\newcommand{\VcSupImgP}{.62}
\newcommand{\VcSupImgCondChi}{8{,}872}
\newcommand{\VcSupDmeanCondChi}{20{,}596}

\newcommand{\VcPcDmeanB}{\ensuremath{-}2.86}
\newcommand{\VcPcDmeandAIC}{\ensuremath{-}153}
\newcommand{\VcSupDminCondChi}{18{,}198}

\newcommand{\VcPcDminB}{\ensuremath{-}2.84}
\newcommand{\VcPcDmindAIC}{+1562}

\newcommand{\VcPosthocDate}{24 November 2026}
\newcommand{\VcDispSub}{.24}
\newcommand{\VcDispSup}{.86}

\newcommand{\VcSupMeanSub}{.89}
\newcommand{\VcSupSdSub}{.10}
\newcommand{\VcSupMeanSup}{.84}
\newcommand{\VcSupSdSup}{.12}

\newcommand{\VdAgreeWord}{eight}
\newcommand{\VdDifferWord}{four}

\newcommand{\VfSupDmeanCondChiMinAll}{14{,}755}

\pretitle{%
  \begin{center}
  \rule{\linewidth}{0.8pt}\par\vspace{0.4em}
  \LARGE\bfseries
}

\posttitle{%
  \par\vspace{0.4em}\rule{\linewidth}{0.8pt}
  \end{center}
}

\preauthor{%
  \begin{center}\normalsize
}

\postauthor{%
  \par\end{center}\vspace{-1.0em}
}

\predate{}
\postdate{}

\title{Visual Similarity in Adult Word Extension and Artificial Word Learning}

\author{
P.~M.~Vera\\
Neurocreaciones\\
\texttt{patricio@neurocreaciones.ai}
}

\date{}

\begin{document}

\maketitle
\vspace{-1.0em}

\begin{abstract}
When two objects share a name, how much does their visual similarity explain, and what do the teaching examples add? We evaluated the limits of visual similarity as an account of word extension across ten visual representations: eight trained encoders that differ in architecture, training data, and objective, a randomly initialized network, and raw pixels. First, we reanalyzed trial-level data from a large adult word-extension study \parencite{lewis2018}, computing each test image's distance to the teaching examples in each representation (\VtPersons{} people). Within the taught domain, greater distance was associated with less extension for all eight trained encoders (log odds per standard deviation from \VtPcViTB{} to \VtPcMaeB) and also for the untrained network (\VtPcRandB), so the association cannot be attributed only to what the encoders learned; raw pixels showed no negative association (\VtPcPixB). In post hoc models that let the distance slope vary across images, \VcImgTrSameSignWord{} of the eight trained encoders retained a negative estimate, and \VcImgTrExclWord{} had 95\% intervals entirely below zero; the interval of the reference encoder included zero. The original study found the suspicious-coincidence effect to depend on whether the one-example trial came before or after the three-example trials, an order that varied between its experiments; whether this order also changes the distance slope was therefore the frozen primary question, tested at the level of the experiments, and it remained inconclusive ($\hat\Delta = \VtOrdDelta$, 95\% CI [\VtOrdLo,~\VtOrdHi]). The composition of the examples, by contrast, changed extension at comparable distances: after three examples, adults extended a word to \SelSubSup\% of superordinate items when the examples were subordinate and to \SelSupSup\% when they were superordinate. Second, in artificial learners built on each representation, words joining more distant categories were learned less accurately for \VtDSupportWord{} of the eight trained encoders. Every trained encoder's geometry correlated with human similarity measures from three datasets, and so did the untrained network's, although far more weakly on the two similarity datasets (Spearman $\rho = \VtHoneRandThings$ and \VtHoneRandHsj, against \VcHoneTrMin{} to \VcHoneTrMax{} for the trained encoders other than MAE); the frozen support rule, which required only a positive correlation, did not separate them. In an exploratory comparison in which adults and learners judged the same images, no common acceptance threshold on any trained encoder reproduced the adults' gain from narrow to broad teaching (at most \VtXJumpMax{} vs.\ \VtXJumpAdult{} percentage points; with 12 candidate images, a gap of about one image), and two learners with equal within-domain acceptance extended the word to different images. Cosine distance to the centroid of the teaching examples, computed on final-layer global embeddings, predicts part of word extension across architectures, even without training, but does not capture the effect of example composition; in post hoc checks, distances to the individual examples did not capture it either. Analyses followed internal plans fixed before the confirmatory runs (not externally preregistered).
\end{abstract}

\vspace{-0.6em}
\noindent\textbf{Keywords:} word learning; word extension; suspicious coincidence; visual similarity; vision encoders; self-supervised vision models; secondary data analysis; artificial learners

\section{Introduction}

Visual similarity is the most obvious candidate explanation for why a novel word spreads from its examples to some objects and not to others. This article evaluates how far that explanation goes. We measured visual similarity with ten representations, eight trained vision encoders that differ in architecture, training data, and objective, plus a randomly initialized network and raw pixels, and asked, for adults and for artificial learners, what distance in each space predicts about word extension and what it leaves out. The pattern is consistent across architectures: distance to the centroid of the teaching examples, measured in each encoder's final-layer global embedding, is associated with part of adult extension, even when it is measured in a network that was never trained, but it does not capture the change in extension produced by the composition of the teaching examples, and the frozen primary test of whether presentation order changes the distance slope was inconclusive. Our conclusions concern this readout of the encoders' geometry, not visual geometry in general: readouts that use the individual examples, their dispersion, or intermediate layers could behave differently, and we examine exemplar-based distances only post hoc.

A word can cover several kinds of objects and still be coherent. Children call cats and dogs \emph{doggie} long before they call frogs and trucks by a common name \parencite{rescorla1980}, and adults told that three Dalmatians are all \emph{feps} will typically refuse the label for a cow. Two kinds of explanation are usually offered for such patterns. Similarity-based accounts hold that a word spreads to referents that resemble its known instances \parencite{sloutsky2004,nosofsky1986}. Bayesian accounts hold that learners infer the most plausible extension from a structured hypothesis space, so that the \emph{composition} of the examples, and not only their similarity to a test object, determines how far the word is extended \parencite{xu2007}. Developmental work shows both influences: children extend novel nouns by shape \parencite{landau1988}, construe them taxonomically rather than thematically \parencite{markman1984}, and project properties by category membership over appearance \parencite{gelman1986}.

This article asks a narrow version of this question: \emph{when two objects receive the same name, how much is explained by their visual similarity, and what do the examples used to teach the word contribute?} Answering it requires a measure of visual similarity that is defined independently of the word-learning behavior under study and can be applied to the exact images people saw. Vision models trained with self-supervision, language supervision, or labels now provide such measures. Their embedding spaces can be queried for any image, and their geometry partly aligns with human similarity judgments \parencite{hebart2020,roads2021}. They are not models of human perception, and human similarity judgments themselves mix perceptual and conceptual information \parencite{hebart2020,hebart2023}; but they give an explicit, reproducible similarity space in which both human and artificial generalization can be described.

A growing literature asks how far such representations align with human similarity and how alignment can be increased. Alignment with human similarity judgments varies across networks and depends more on training data and objective than on model scale or architecture \parencite{muttenthaler2023}. Representations can be made more human-like by fine-tuning on human perceptual similarity judgments \parencite{fu2023}, which also benefits several downstream vision tasks \parencite{sundaram2024}, or by transferring the hierarchical structure of human judgments across abstraction levels \parencite{muttenthaler2025}. We deliberately used frozen encoders that were not aligned to human data, so that any convergence with human measures is not built in, and we compared encoders that differ in training data and objective with two controls that have no learned features: a randomly initialized network and raw pixels. The partial alignment we report should be read against these choices. We used one global embedding per image from the final layer of each encoder (Appendix~\ref{app:checkpoints}); intermediate layers and patch-level features may relate differently to human similarity and were not examined.

We combine two sources of evidence. The first is human: we reanalyzed trial-level data from \textcite{lewis2018}, who conducted a large, multi-experiment replication of the suspicious-coincidence paradigm of \textcite{xu2007} with adults and made their data, stimuli, and code public. Because the exact images are available, we could compute, for every test item, its distance in each encoder's space to the examples a participant had just seen, and ask whether that distance is associated with the participant's decision to extend the word. The second is artificial: a learner that operates over each frozen encoder, for which we fix which categories each word designates and measure whether the learner can name new images. The artificial learner is not a model of the adult participant; it is a controlled system in which the representation is fully known, so that the contribution of similarity in that space can be isolated. Finally, in an exploratory comparison, we gave artificial learners the same teaching and test images that adults saw.

Presentation order is the primary question of the multi-encoder analysis for a specific reason. \textcite{lewis2018} found that the suspicious-coincidence effect depended on whether the one-example trial came before or after the three-example trials, an order that varied between their 12 experiments. In an exploratory reanalysis with the reference encoder (Section~\ref{sec:results}), a model fitted to individual decisions suggested that order also changed the slope of visual distance. That model treats \OrdNobs{} decisions as independent evidence about a factor that varied across only 12 experiments. The frozen multi-encoder plan therefore made order moderation its primary question and tested it at the experiment level, asking whether the result survived inference at the level at which order was assigned and whether it depended on which encoder defined distance; the association of distance with extension, already supported for the reference encoder, served as a positive control.

The two parts address different behaviors. The adult decided how far to extend a word; the learner tried to recover a taught label. We therefore treat the comparison as a test of how far a shared similarity space describes both, not as evidence of a common mechanism. Our contributions are:
\begin{itemize}
\item A secondary analysis of adult word extension across ten visual representations: within the taught domain, distance is associated with extension for every trained encoder and for an untrained network; the frozen primary test of order moderation is inconclusive; and, in the reference encoder, the effect of distance depends on the composition of the teaching examples (exploratory).
\item A controlled demonstration, in artificial learners, that the learnability of a two-category word decreases with the distance between its categories for most trained encoders, with taxonomic relatedness adding no detectable prediction in the sample tested.
\item Tests of two links between the systems: every trained encoder's geometry, and also that of the untrained network, partially aligns with human similarity and confusability measures, but this alignment did not translate into conclusive evidence that human-derived distances predict which words the learner finds hard.
\item An exploratory comparison on the same images showing that equal within-domain acceptance can hide different extension boundaries, and that no common threshold on a trained encoder reproduced the adults' response to example composition.
\end{itemize}

All human data are secondary: they come from experiments designed and run by other teams to answer other questions.

\section{Adults: Visual Distance and the Composition of Examples}
\label{sec:adults}

\subsection{Data}

\textcite{lewis2018} ran 12 online experiments with adults in the word-learning paradigm of \textcite{xu2007}. On each trial a novel word was taught with one subordinate example, three subordinate examples, three basic-level examples, or three superordinate examples (e.g., one Dalmatian; three Dalmatians; three dogs of different breeds; three different animals), and participants selected, from an array of 24 test images, all the objects that the word could also name. Of the 24 test items, 8 belonged to the taught domain (subordinate, basic-level, and superordinate matches); the rest came from other domains. The suspicious-coincidence effect is the reduction in basic-level generalization after three subordinate examples relative to one. \textcite{lewis2018} recovered the effect with a large effect size under both sequential and simultaneous presentation, but found it sensitive to trial order: it was absent, or reversed under sequential presentation, when the three-example trials preceded the one-example trial. We used their public, anonymized trial-level data and the original stimulus images. Participant identifiers in that release are assigned across experiments, so the data comprise \HFourSessions{} sessions from \HFourPersons{} distinct people (\HFourRepeaters{} people took part more than once), \HFourTrials{} trials, and \HFourNobs{} item-level decisions.

\subsection{Visual Representations and Distance}

We measured visual distance with ten representations. The eight trained encoders were DINOv2-small and DINOv2-base \parencite{oquab2023}; the vision encoder of CVCL, trained on one child's head-camera recordings \parencite{vong2024}; SigLIP \parencite{zhai2023}; an OpenCLIP ViT-B/32 model trained on LAION-2B \parencite{cherti2023,schuhmann2022}; a supervised ViT-B/16 \parencite{dosovitskiy2020} and a supervised ResNet-50 \parencite{he2016}, both trained on ImageNet \parencite{deng2009}; and a masked autoencoder \parencite[MAE;][]{he2022}. The two controls were a randomly initialized ResNet-50, which has the architecture but no training, and raw pixels. Checkpoints, pinned revisions, and the embedding taken from each model are listed in Appendix~\ref{app:checkpoints}; for brevity we refer to each by these names. DINOv2-small, the encoder also used for the artificial learners in Section~\ref{sec:learners}, is the reference: we report its analyses in detail and use the other representations to test how far the results depend on it. For every trial and test item we computed $d$, the cosine distance between the item's L2-normalized embedding and the normalized centroid of the training examples shown on that trial, separately in each representation. The distance is defined entirely by the images, independently of participants' responses. Because the centroid is renormalized, $d$ does not use how dispersed the teaching examples are; Appendix~\ref{app:posthoc} compares it post hoc with distances to the individual examples.

\subsection{Analysis}

\paragraph{Analysis plans.} All analyses followed internally frozen plans (not externally preregistered); deviations and timestamps are documented in Appendix~\ref{app:deviations} and the repository. The DINOv2-small analyses of the human data followed the first plan; the multi-encoder adult and learner analyses followed later plans, frozen before those runs, in which results already known for DINOv2-small and CVCL were treated as descriptive. The comparison on the same images (Section~\ref{sec:matched}) was exploratory and was not frozen.

For the reference encoder, the confirmatory hypothesis (H4 in the analysis plan) was that the probability of selecting a test item decreases with $d$, controlling for the item's taxonomic level and the teaching condition. We fitted a logistic mixed model, $\mathrm{selected} \sim z(d) + \mathrm{level} + \mathrm{condition} + (1 + z(d) \mid \mathrm{person}) + (1 \mid \mathrm{image})$, over all 24 items. The level is the item's taxonomic level within its own domain (subordinate, basic, or superordinate, from the stimulus file name), so out-of-domain items keep their within-domain level; the image factor indexes the 24 stimulus slots (e.g., \texttt{c3\_sup2}), each a single photograph used in all experiments (12 levels in the superordinate-only model). Random effects are grouped by person, as the plan specified (Appendix~\ref{app:deviations}). Support required a negative coefficient with Holm-adjusted $p < .05$ within the six-test confirmatory family. Distance was standardized within each model, so a standard deviation corresponds to \SDdAll{} cosine units over all 24 items and to \SDdTarget{} over the 8 target-domain items; coefficients from the two item sets are therefore not on exactly the same scale. Sensitivity analyses used only the 8 target-domain items, only each person's first session (identified by its recorded start time), excluded people who participated more than once, or excluded the two experiments that were not preregistered by the original authors. The plan listed a distance $\times$ condition interaction as exploratory. A descriptive comparison of conditions and a post hoc test of the incremental contribution of distance were not part of the plan and are reported as exploratory. Two further post hoc exploratory analyses were specified after the main results were known, and before they were run (Appendix~\ref{app:deviations}). First, a superordinate-only model restricted to the four superordinate target-domain items of each trial (12 images) compared $\mathrm{selected} \sim z(d) + \mathrm{condition}$ with models dropping either term, with random intercepts by person and image (sensitivity: adding a by-person distance slope). Second, a trial-order model used the experiment-level order of the one-example and three-example trials (one-example trial first or last; six experiments each) and tested, by likelihood ratio, whether order moderated the distance slope ($z(d) \times \mathrm{order}$, 1 \textit{df}) and the distance $\times$ condition interaction ($z(d) \times \mathrm{condition} \times \mathrm{order}$, 3 \textit{df}), with Holm correction over these two tests. Order varies only between experiments and is therefore confounded with other experiment-level differences.

\paragraph{Distance and order across representations.} Visual distance and presentation order are separate questions, and the multi-encoder analyses address them separately. The \emph{positive control} asks whether distance is associated with extension: for each representation, the model above was fitted to the eight within-domain test images (\VtPcNobs{} observations, \VtPersons{} people). In the frozen plan it was a gate, not the primary test. The \emph{primary} analysis asks whether presentation order changes the distance slope. Order varied across the \VtNexp{} experiments, so inference was at the experiment level: a two-stage permutation test of the slope difference $\hat\Delta$, classified into decision regions at three effect margins ($\delta = .05, .10, .20$) and summarized by a global rule across margins and geometry cutoffs; Appendix~\ref{app:rules} gives the formal definitions. Distance was standardized within each representation and item set, so each slope is in log odds per standard deviation of that representation's own distance (raw standard deviations in Appendix Table~\ref{tab:posthoc}). We also refitted the 24-item H4 model with each representation (\VtHfourNobs{} observations). Mixed models were fitted with lme4 \parencite{bates2015}.

\subsection{Results}
\label{sec:results}

\begin{figure}[!htbp]
\centering
\includegraphics[width=\linewidth]{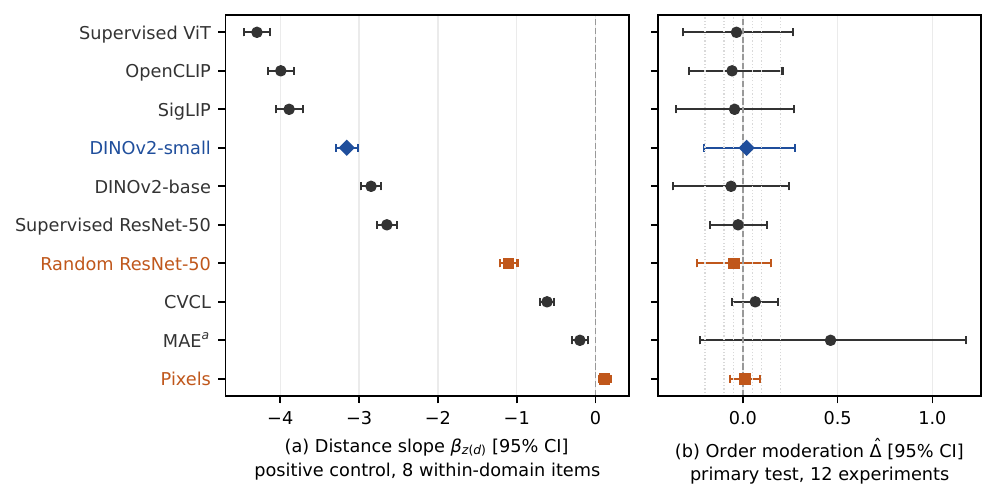}
\caption{Adult Word Extension and Visual Distance by Representation}
\label{fig:forest}
\par\smallskip{\raggedright\footnotesize\textit{Note.} (a) Positive control: distance slope (log odds per standard deviation of $d$) in the logistic mixed model fitted to the eight within-domain test images (\VtPcNobs{} observations, \VtPersons{} adults), with 95\% Wald intervals; rows are ordered by this slope. Blue: reference encoder; black: trained encoders; orange: controls. $^{a}$MAE: fitted with random intercepts only, because the model with by-person distance slopes did not pass the convergence and non-singularity check. (b) Primary test: experiment-level slope difference by presentation order over \VtNexp{} experiments, with 95\% intervals; dotted lines mark the effect margins $\pm.05$, $\pm.10$, and $\pm.20$. The frozen global rule was inconclusive. Exact values, the 24-item model, and the decision regions are in Appendix Table~\ref{tab:v2adults}.\par}
\end{figure}

\paragraph{Distance across representations.} Figure~\ref{fig:forest}a shows the positive-control slopes. All \VtPcTrainedNeg{} trained encoders gave a negative distance slope whose 95\% interval excluded zero, from \VtPcViTB{} [\VtPcViTLo,~\VtPcViTHi] for supervised ViT to \VtPcMaeB{} [\VtPcMaeLo,~\VtPcMaeHi] for MAE; the reference, DINOv2-small, gave \VtPcRefB{} [\VtPcRefLo,~\VtPcRefHi]. The randomly initialized ResNet-50 also gave a negative slope, \VtPcRandB{} [\VtPcRandLo,~\VtPcRandHi], steeper than those of CVCL (\VbPcCvclB{} [\VbPcCvclLo,~\VbPcCvclHi]) and MAE, whereas pixels gave \VtPcPixB{} [\VtPcPixLo,~\VtPcPixHi]. Because each slope is per standard deviation of that representation's distance, and the MAE model omits the by-person distance slope (omitting a random slope tends to pull a conditional logit slope toward zero), the ranking of slopes across representations is approximate. The association therefore appears with several representations, including one with no training, and cannot be attributed only to what the encoders learned. The primary question was not resolved (Figure~\ref{fig:forest}b): for DINOv2-small the slope difference by order was $\hat\Delta = \VtOrdDelta$ (95\% CI [\VtOrdLo,~\VtOrdHi]), and the frozen global rule returned \VtGlobalLabel{} (inconclusive) at all \VtGlobalCells{} combinations of margin and geometry cutoff. Because the reference encoder was inconclusive, we make no claim that order moderation is equivalent across encoders. In the 24-item H4 model, all \VbHfourAllNeg{} representations, including both controls, gave negative slopes, and their ordering differs from that of the positive control (e.g., \VbHfourRandB{} for the untrained network, \VbHfourCvclB{} for CVCL, and \VbHfourMaeB{} for MAE; Appendix Table~\ref{tab:v2adults}); as for DINOv2-small below, these coefficients are dominated by the contrast between the taught domain and the other domains.

\paragraph{The reference encoder in detail.} Visual distance was associated with extension: the log odds of selecting an item fell by \HFourB{} per standard deviation of distance (95\% CI [\HFourLo,~\HFourHi]; Holm-adjusted $p < .001$), with level and condition in the model. The estimate was stable across specifications (Table~\ref{tab:h4}): \HFourTargetB{} with only the 8 target-domain items, \HFourFirstB{} with first sessions only, \HFourExclB{} excluding repeat participants, and \HFourTargetExclB{} combining the last restriction with target items. Over all 24 items, however, the coefficient is dominated by the contrast between the taught domain and the other domains, whose items were almost never selected. The target-only analyses address this. In an exploratory post hoc comparison restricted to the 8 target-domain items, adding distance to a model with level and condition (and the same random intercepts) improved fit markedly (likelihood ratio $\chi^2(1) = \IncChi$, $\Delta\mathrm{AIC} = \IncDAIC$; $\chi^2(3) = \IncChiRE$ when the by-person distance slope was also added). Within each taxonomic level the distance slope was negative: \LvlSubB{} [\LvlSubLo,~\LvlSubHi] for subordinate, \LvlBasB{} [\LvlBasLo,~\LvlBasHi] for basic-level, and \LvlSupB{} [\LvlSupLo,~\LvlSupHi] for superordinate items (log odds per target-model standard deviation). Distance and level remain correlated by construction in this stimulus set, so these estimates describe an association beyond the coded level, not a separable causal contribution of similarity.

\begin{table}[!htbp]
\centering\small
\begin{tabular}{lcrl}
\toprule
Specification (random effects by person) & People & \multicolumn{2}{c}{Log odds per \textit{SD} of $d$} \\ \cmidrule(lr){3-4}
 & & \multicolumn{1}{c}{Est.} & \multicolumn{1}{c}{95\% CI} \\
\midrule
Primary: all 24 items & \HFourPersons & \HFourBshort{} & \mbox{[\HFourLoshort,~\HFourHishort]} \\
First session only & \HFourPersons & \HFourFirstB{} & \mbox{[\HFourFirstLo,~\HFourFirstHi]} \\
Excluding repeat participants & \HFourExclPersons & \HFourExclB{} & \mbox{[\HFourExclLo,~\HFourExclHi]} \\
Excluding non-preregistered experiments & \HFourPreregPersons & \HFourPreregB{} & \mbox{[\HFourPreregLo,~\HFourPreregHi]} \\
8 target-domain items & \HFourPersons & \HFourTargetB{} & \mbox{[\HFourTargetLo,~\HFourTargetHi]} \\
8 target items, excluding repeat participants & \HFourExclPersons & \HFourTargetExclB{} & \mbox{[\HFourTargetExclLo,~\HFourTargetExclHi]} \\
\bottomrule
\end{tabular}
\caption{Association Between Visual Distance and Word Extension in Adults (H4)}
\label{tab:h4}

\par\smallskip{\raggedright\footnotesize\textit{Note.}  Logistic mixed models with level and condition as fixed effects and by-person random intercepts and distance slopes plus by-image random intercepts; Wald intervals. All $p < .001$.\par}
\end{table}

\begin{figure}[!htbp]
\centering
\includegraphics[width=\linewidth]{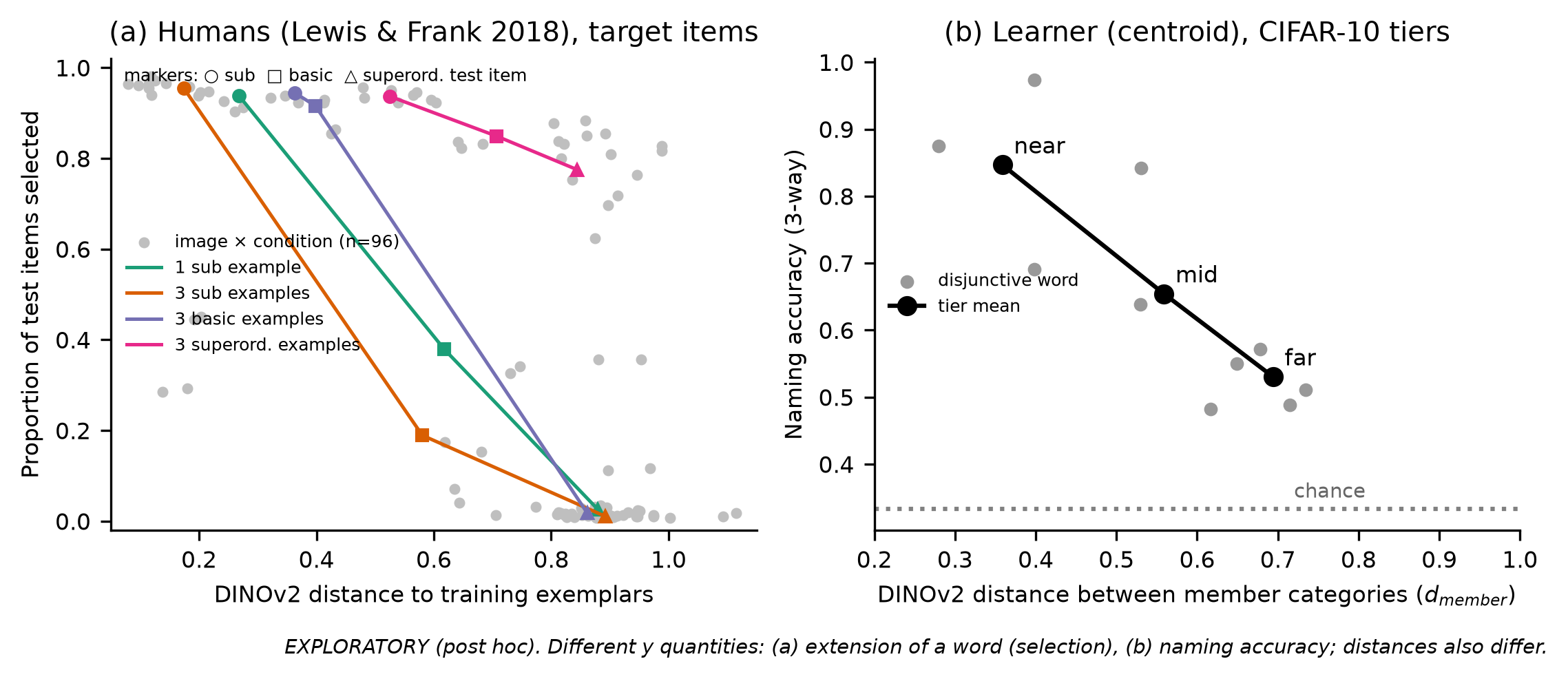}
\caption{Visual Distance and Generalization in Adults and in the Artificial Learner. \emph{Note.} Exploratory comparison. (a) Proportion of target-domain test items selected by adults in \textcite{lewis2018}, by teaching condition and test level, against DINOv2-small distance to the training examples; gray points are the \XUnits{} image $\times$ condition units (descriptive logit slope \XSlope{} [\XSlopeLo,~\XSlopeHi], $\rho = \XRho$). (b) Naming accuracy of the centroid learner for two-category words in CIFAR-10, against the DINOv2-small distance between member categories. The vertical axes measure different quantities (extension vs.\ naming accuracy) and should not be compared directly.}
\label{fig:x1}
\end{figure}

Distance, however, was far from the whole story. Figure~\ref{fig:x1}a shows mean selection for the target-domain items by teaching condition. For superordinate test items, selection was low after one subordinate example (\SelOneSup\%), three subordinate examples (\SelSubSup\%), or three basic-level examples (\SelBasSup\%), but rose to \SelSupSup\% after three superordinate examples. Mean distances of these items to the examples differed only modestly across conditions (range \DistSupMin--\DistSupMax: \DistOneSup, \DistSubSup, \DistBasSup, and \DistSupSup, respectively; distributions in Appendix Figure~\ref{fig:supdist}). The number of examples was the same in the last three conditions; what changed was their composition. The overall mean for superordinate items (\SelSupOverall\%) conceals this contrast. These percentages are descriptive.

In an exploratory model with a distance $\times$ condition interaction (random effects by person; all 24 items), the distance slope was \IntRefB{} [\IntRefLo,~\IntRefHi] after one example and changed little after three subordinate examples (interaction \IntSubB{} [\IntSubLo,~\IntSubHi]), but became steeper after three basic-level (\IntBasB{} [\IntBasLo,~\IntBasHi]) and three superordinate examples (\IntSupB{} [\IntSupLo,~\IntSupHi]); the interaction improved fit substantially (likelihood ratio $\chi^2(3) = \IntChi$, $p < .001$). Thus, how strongly visual distance relates to extension depends on how the word was taught. Note that this model uses all 24 items, whereas the curves in Figure~\ref{fig:x1}a show only the 8 target-domain items. Its condition-specific slopes are therefore driven largely by the contrast between the taught domain and the other domains, whose items stayed near zero while within-domain selection rose after broad teaching; they are not the slopes of the within-domain curves in Figure~\ref{fig:x1}a.

In the post hoc superordinate-only model (\SupNobs{} observations, \SupPersons{} people, \SupImages{} images), condition improved fit given distance ($\chi^2(3) = \SupCondChi$, $p \SupCondP$), and distance improved fit given condition ($\chi^2(1) = \SupDistChi$, $p \SupDistP$), with a distance coefficient of \SupZdB{} [\SupZdLo,~\SupZdHi] per within-subset standard deviation. Relative to one example, the condition coefficients were \SupCSubB{} [\SupCSubLo,~\SupCSubHi] for three subordinate, \SupCBasB{} [\SupCBasLo,~\SupCBasHi] for three basic-level, and \SupCSupB{} [\SupCSupLo,~\SupCSupHi] for three superordinate examples. Adding a by-person distance slope gave similar estimates (distance \SupSlZdB{} [\SupSlZdLo,~\SupSlZdHi]; the distance-only model in this sensitivity analysis was singular). In the post hoc trial-order model (\OrdNobs{} observations, \OrdPersons{} people), the distance $\times$ order term was \OrdZdB{} [\OrdZdLo,~\OrdZdHi] ($\chi^2(1) = \OrdZdChi$, $p \OrdZdP$, Holm $p \OrdZdPHolm$), and the three-way distance $\times$ condition $\times$ order terms gave $\chi^2(3) = \OrdIntChi$ ($p \OrdIntP$, Holm $p \OrdIntPHolm$). With the three-way terms in the model, the distance $\times$ order term (for the one-example reference condition) fell to \OrdZdTriB{} ($p \OrdZdTriP$), so moderation of the distance slope by order depended on the teaching condition. These likelihood-ratio tests treat individual decisions as independent evidence about order, which varied only between the 12 experiments, and the model has no experiment-level random effects; the experiment-level primary test, designed for this problem, was inconclusive, and the post hoc $p$ values do not override it.

\paragraph{Post hoc checks after review (exploratory).} Three checks were added after the analyses and two automated external reviews were complete (Appendix~\ref{app:posthoc}). Adding by-image random slopes for distance to the positive-control model (all ten fits converged) kept the slope negative for \VcImgTrSameSignWord{} of the eight trained encoders but widened every interval: it excluded zero for \VcImgTrExclWord{} and included it for \VcImgTrInclList{} (DINOv2-small: \VcImgRzB{} [\VcImgRzLo,~\VcImgRzHi], with a standard deviation of \VcImgRzSd{} for the image slopes). Under this model the reference encoder would not have passed the positive control; the random network kept an interval below zero (\VcImgRandB{} [\VcImgRandLo,~\VcImgRandHi]). In the superordinate-only model, distance no longer improved fit given condition once its slope could vary across images ($\chi^2(1) = \VcSupImgChi$, $p = \VcSupImgP$). Models with an intercept for each image--condition cell did not converge for any representation. Distances to the individual teaching examples did not absorb the effect of composition: in the superordinate-only model, teaching condition still improved fit given the mean-exemplar distance ($\chi^2(3) = \VcSupDmeanCondChi$ for DINOv2-small; at least \VfSupDmeanCondChiMinAll{} for every representation), and given the nearest-exemplar distance ($\chi^2(3) = \VcSupDminCondChi$). Across the eight within-domain items, the mean-exemplar distance fitted better than the centroid distance for \VcDmeanBetterN{} of the ten representations and the nearest-exemplar distance for \VcDminBetterN{} (AIC).

\section{Artificial Learners: A Controlled System}
\label{sec:learners}

The adult data show an association but cannot isolate the role of similarity, because the experimenter does not control the representation the participant uses. In an artificial learner the representation is known and the extension of each word can be fixed. We summarize the core results of the artificial-learner study (full details of the original experiments are in \textcite{vera2026lexical} and in the project repository).

\subsection{Method}

Images from CIFAR-10 and CIFAR-100 \parencite{krizhevsky2009} were encoded once with frozen DINOv2-small \parencite{oquab2023} (384-dimensional CLS embeddings, L2-normalized). For DINOv2-small, images were preprocessed with the model's default Hugging Face image processor: conversion to RGB, bicubic resizing of the shorter side to 256 pixels (the 32 $\times$ 32 CIFAR images are therefore upsampled), a 224 $\times$ 224 center crop, rescaling to [0, 1], and normalization with ImageNet means and standard deviations. The same preprocessing was applied to the stimulus images of the adult study. The other representations used their own preprocessing (Appendix~\ref{app:checkpoints}), which on square CIFAR images keeps either the full frame or its central 87.5\% per side. We did not test alternative preprocessing choices (e.g., other interpolation methods or crops). Learners received nonce words (e.g., \emph{slithy}, \emph{mimsy}, \emph{vorpal}; \cite{carroll1871}) paired with five images each, in 3-way episodes, and were tested on naming held-out images (naming accuracy, macro-averaged over words, is labeled C1 in figures and tables). The main learner represents a word by the normalized mean of its training embeddings (a prototype; equivalent to a prototypical network over a frozen encoder; \cite{snell2017}); exemplar ($k$-nearest-neighbor) and logistic-regression learners gave the same pattern. Each word was assigned either one category (native) or two categories (disjunctive). Disjunctive words were grouped into near, mid, and far tiers by the cosine distance between the two categories' centroids (mean \DmemNear, \DmemMid, and \DmemFar). For example, \emph{tromble} (cat + dog) and \emph{mivor} (frog + truck) both join exactly two categories; what differs is the distance between them. Thirty paired episodes per condition were run (seed 42), with 95\% bootstrap intervals over episodes. The tiers contain few words (\NPairsNear, \NPairsMid, and \NPairsFar{} category pairs for near, mid, and far), so these intervals reflect episode-level variability and do not generalize over words; the dissociation study below addresses generalization over pairs.

To separate distance in the encoder's space from taxonomic relatedness, a dissociation study on CIFAR-100 sampled 100 category pairs from the four quadrants formed by median splits of perceptual (DINOv2) distance and semantic distance (one minus WordNet Wu--Palmer similarity; \cite{wu1994,miller1995}), taught each pair as one word, and regressed naming accuracy on both distances, under decision rules specified in advance in an internal plan. The reported dissociation results use the original, pre-specified mapping of class names to WordNet synsets. That mapping selected an unintended sense for \WnNCorrected{} classes (e.g., \emph{tiger} as a fierce person; \WnNTouch{} of the 100 sampled pairs involve at least one of them); with these senses corrected, the conclusion was unchanged (perceptual slope $\beta = \WnCorrBeta$ and semantic-distance $p = \WnCorrPsem$, versus $\beta = \WnFrozenBeta$ and $p = \WnFrozenPsem$ with the original mapping; with quadrants reassigned, median accuracy remained higher in Q4 than in Q3, \WnCorrQFour{} vs.\ \WnCorrQThree, $n = \WnCorrNQ$ discordant pairs). Details are in the released materials.

\paragraph{Across representations.} The same learners were run on embeddings from each of the ten representations (Appendix~\ref{app:checkpoints}). For each we report naming accuracy by tier (T), the partial correlation between inter-category distance and naming accuracy for the 100 pairs of the dissociation study, adjusting for WordNet distance (D), and the same analysis on all \VtDallN{} CIFAR-100 pairs (D-all; a sensitivity analysis, since the pairs share categories). The same 100 pairs, selected with DINOv2-small distances, were used for every representation, each with its own distances, so the quadrants are balanced only for DINOv2-small. Results were classified into decision regions at smallest effect sizes of interest (SESOI) of .10, .20, and .30 \parencite{schuirmann1987,lakens2017}, with the rules of the frozen plan (Appendix~\ref{app:rules}): support requires an effect in the predicted direction with unadjusted $p < .05$ and does not depend on the SESOI, which enters only the no-support region. Because the DINOv2-small and CVCL results were known before the plan was frozen, they are descriptive and outside the correction families; the other six trained encoders are confirmatory, and the two controls are floors. Distance effects in T and D are close to mechanical for a learner that operates on the encoder's distances, so only the links tested in Section~\ref{sec:link} are read against people; T and D characterize each representation as a substrate for the learner (how separable its categories are for a prototype learner) rather than test a claim about people.

\subsection{Results}

\begin{figure}[!htbp]
\centering
\includegraphics[width=0.75\linewidth]{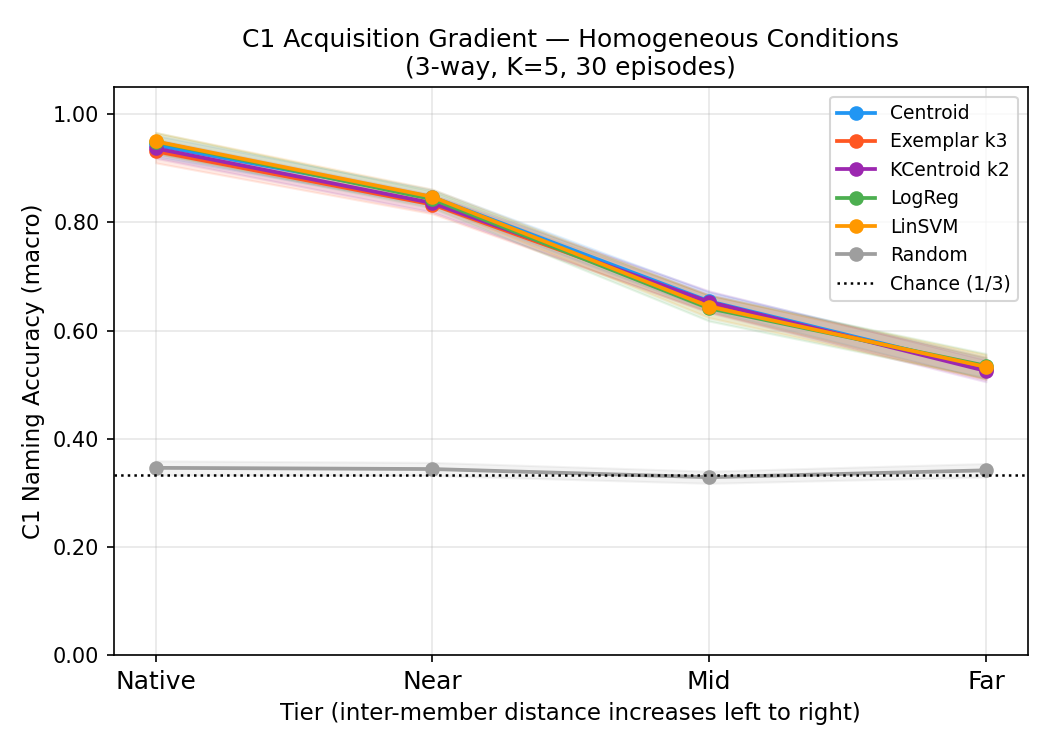}
\caption{Naming Accuracy of Artificial Learners Across Concept Tiers. \emph{Note.} 3-way, 5-shot episodes; shaded bands are 95\% bootstrap intervals over 30 episodes. All non-random learners overlap; the random baseline stays at chance.}
\label{fig:c1}
\end{figure}

\begin{figure}[!htbp]
\centering
\includegraphics[width=0.65\linewidth]{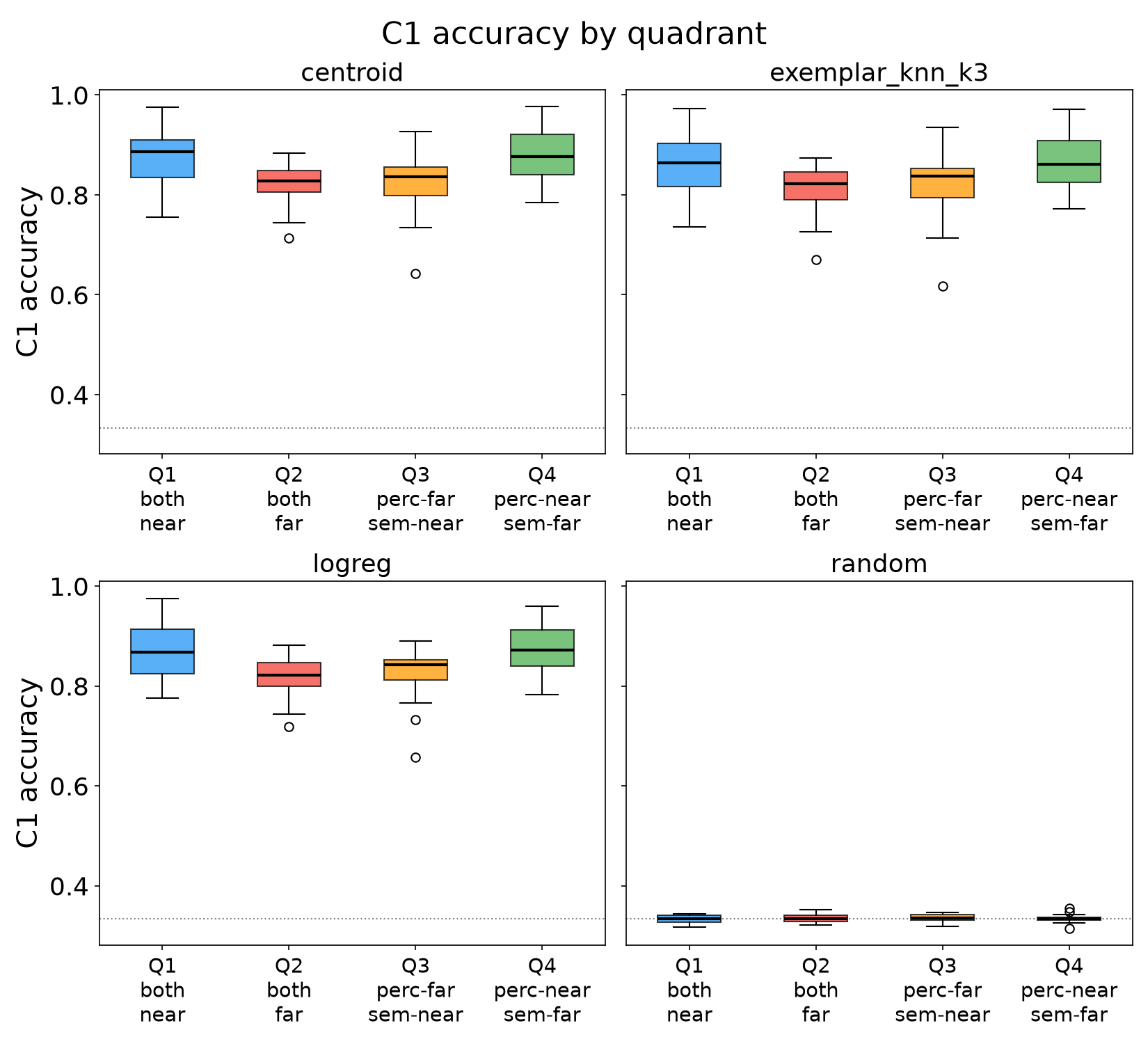}
\caption{Naming Accuracy by Dissociation Quadrant (CIFAR-100). \emph{Note.} Quadrants are defined by median splits on perceptual (DINOv2 inter-centroid cosine) and semantic (WordNet Wu--Palmer) distance. Accuracy follows the perceptual axis.}
\label{fig:quad}
\end{figure}

Naming accuracy was \AccNative\% [\AccNativeLo,~\AccNativeHi] for native words and fell to \AccNear\% [\AccNearLo,~\AccNearHi], \AccMid\% [\AccMidLo,~\AccMidHi], and \AccFar\% [\AccFarLo,~\AccFarHi] for near, mid, and far disjunctive words (chance $\approx 33\%$; Figure~\ref{fig:c1}). The ordering was the same for every non-random learner and replicated across five master seeds. In the dissociation study, perceptual distance predicted accuracy ($\beta = \DisBeta$, $p = \DisPperc$, partial $R^2 = \DisPRperc$), whereas semantic distance did not ($p = \DisPsem$, partial $R^2 = \DisPRsem$; Figure~\ref{fig:quad}); sharing a CIFAR-100 superclass did not add detectable prediction once perceptual distance was controlled ($\beta = \DisSuperB$, $p = \DisSuperP$). Because the learner has no access to semantic information, this is a sufficiency result: distance in the learner's own space suffices to produce the gradient; it does not show that semantics is irrelevant for human learners. Pairs within the same superclass were, on average, learned more accurately (\AccWithin\% vs.\ \AccBetween\%), but only \NWithin{} of the 100 pairs fell within a superclass.

\paragraph{Across representations.} Table~\ref{tab:agents} summarizes the learners. Greater distance between the two categories joined by a word predicted lower naming accuracy (D, support region at all three SESOI) for \VtDSupportWord{} of the eight trained encoders, with partial correlations from \VtDrMin{} to \VtDrMax{} on the \VtDN{} pairs. CVCL ($r = \VtDCvclR$ [\VtDCvclLo,~\VtDCvclHi]) and MAE ($r = \VtDMaeR$ [\VtDMaeLo,~\VtDMaeHi]) did not reach support; both did in the descriptive D-all analysis, whose pairs share classes and are not independent (its regions treat the \VtDallN{} pairs as independent and are descriptive only). Neither control reached support in D.

\begin{table}[!htbp]
\centering\small
\setlength{\tabcolsep}{3.5pt}
\noindent\adjustbox{max width=\linewidth}{%
\begin{tabular}{llrrrlr}
\toprule
 & & & & \multicolumn{2}{c}{D: partial $r$} & D-all \\ \cmidrule(lr){5-6}
Encoder & Role & Native (\%) & Far (\%) & \multicolumn{1}{c}{Est.} & \multicolumn{1}{c}{95\% CI} & partial $r$ \\
\midrule
DINOv2-small & Descriptive & 94.3 & 53.0 & \ensuremath{-}.50 & \mbox{[\ensuremath{-}.63,~\ensuremath{-}.33]} & \ensuremath{-}.60 \\
DINOv2-base & Confirmatory & 96.3 & 54.4 & \ensuremath{-}.57 & \mbox{[\ensuremath{-}.69,~\ensuremath{-}.42]} & \ensuremath{-}.69 \\
CVCL & Descriptive & 67.5 & 42.8 & .015 & \mbox{[\ensuremath{-}.18,~.21]}$^{\dagger}$ & \ensuremath{-}.09 \\
SigLIP & Confirmatory & 95.8 & 53.4 & \ensuremath{-}.30 & \mbox{[\ensuremath{-}.47,~\ensuremath{-}.11]} & \ensuremath{-}.48 \\
OpenCLIP & Confirmatory & 97.0 & 52.4 & \ensuremath{-}.35 & \mbox{[\ensuremath{-}.51,~\ensuremath{-}.16]} & \ensuremath{-}.52 \\
Supervised ViT & Confirmatory & 97.8 & 51.8 & \ensuremath{-}.65 & \mbox{[\ensuremath{-}.75,~\ensuremath{-}.52]} & \ensuremath{-}.68 \\
Supervised ResNet-50 & Confirmatory & 87.0 & 49.5 & \ensuremath{-}.30 & \mbox{[\ensuremath{-}.47,~\ensuremath{-}.11]} & \ensuremath{-}.39 \\
MAE & Confirmatory & 48.3 & 38.8 & \ensuremath{-}.004 & \mbox{[\ensuremath{-}.20,~.19]}$^{\dagger}$ & \ensuremath{-}.07 \\
\midrule
Random ResNet-50 (s0) & Control & 41.0 & 38.4 & .18 & \mbox{[\ensuremath{-}.02,~.37]}$^{\dagger}$ & .04$^{\dagger}$ \\
Pixels & Control & 45.8 & 38.7 & \ensuremath{-}.14 & \mbox{[\ensuremath{-}.33,~.06]}$^{\dagger}$ & \ensuremath{-}.18 \\
\bottomrule
\end{tabular}}

\caption{Artificial Learners on CIFAR by Representation}
\label{tab:agents}
\par\smallskip{\raggedright\footnotesize\textit{Note.} Native and Far: centroid naming accuracy (\%) for native words and far two-category words in CIFAR-10 (chance $\approx 33\%$). D: partial correlation between inter-category distance and naming accuracy, adjusting for WordNet distance, on the \VtDN{} pairs of the dissociation study; D-all: the same on all \VtDallN{} pairs (descriptive; pairs share classes). $^{\dagger}$Not in the support region at all three SESOI (.10/.20/.30). DINOv2-small and CVCL: descriptive; controls: floors. Intervals for accuracy and the decision regions are in Appendix Table~\ref{tab:v2agentsD}.\par}
\end{table}

\section{Testing the Link Between the Two Systems}
\label{sec:link}

The two results above use the same similarity spaces, but the adult result does not show that these spaces capture human similarity, and the learner result does not show that human similarity predicts learning. We tested these two distinct relations, alignment between similarity measures (H1) and prediction of learner difficulty from human distances (H2), with public human datasets, under the same analysis plan.

\subsection{Data}

We used three independent human datasets that cover CIFAR categories: odd-one-out similarity judgments for THINGS concepts \parencite{hebart2020,hebart2023}, similarity judgments for ImageNet images (ImageNet-HSJ; \cite{roads2021}), and human classification uncertainty for the CIFAR-10 test set (CIFAR-10H; \cite{peterson2019}). Each was converted to a distance between category pairs. CIFAR-100 classes were mapped to THINGS concepts by name and WordNet sense, with ambiguous names resolved to the sense shown in the CIFAR images (e.g., \emph{mouse} as the rodent, \emph{tank} as the military vehicle); \MapThingsClasses{} classes were mapped. Three classes that share a single THINGS concept (maple, oak, and willow, all mapped to \emph{tree}) kept their mapping, but the pairs among them were excluded, which yields \CovThingsPairs{} pairs; \emph{cattle} and \emph{ray} have only hyponym matches in THINGS and were used only in a stricter sensitivity mapping. For ImageNet-HSJ, each CIFAR class was assigned the ImageNet synsets in the hyponym closure of its WordNet synset (synsets claimed by more than one class went to the most specific class, and exact ties were dropped); \MapHsjClasses{} CIFAR-100 classes received at least one synset, and pairs were retained when at least 10 relevant judgments were available, yielding \HOneHsjN{} pairs. CIFAR-10H uses the CIFAR-10 classes directly. Robustness checks with alternative mappings are in the released materials. All mappings were fixed before human distances were computed. THINGS and ImageNet-HSJ judgments reflect both perceptual and conceptual properties of objects \parencite{hebart2020}, and CIFAR-10H measures confusability rather than similarity; none of them is a pure measure of visual similarity. Confirmatory tests (H1 $\times$ 3, H2 $\times$ 2, H4) were Holm-corrected as one family.

\subsection{Results}

\begin{figure}[!htbp]
\centering
\includegraphics[width=\linewidth]{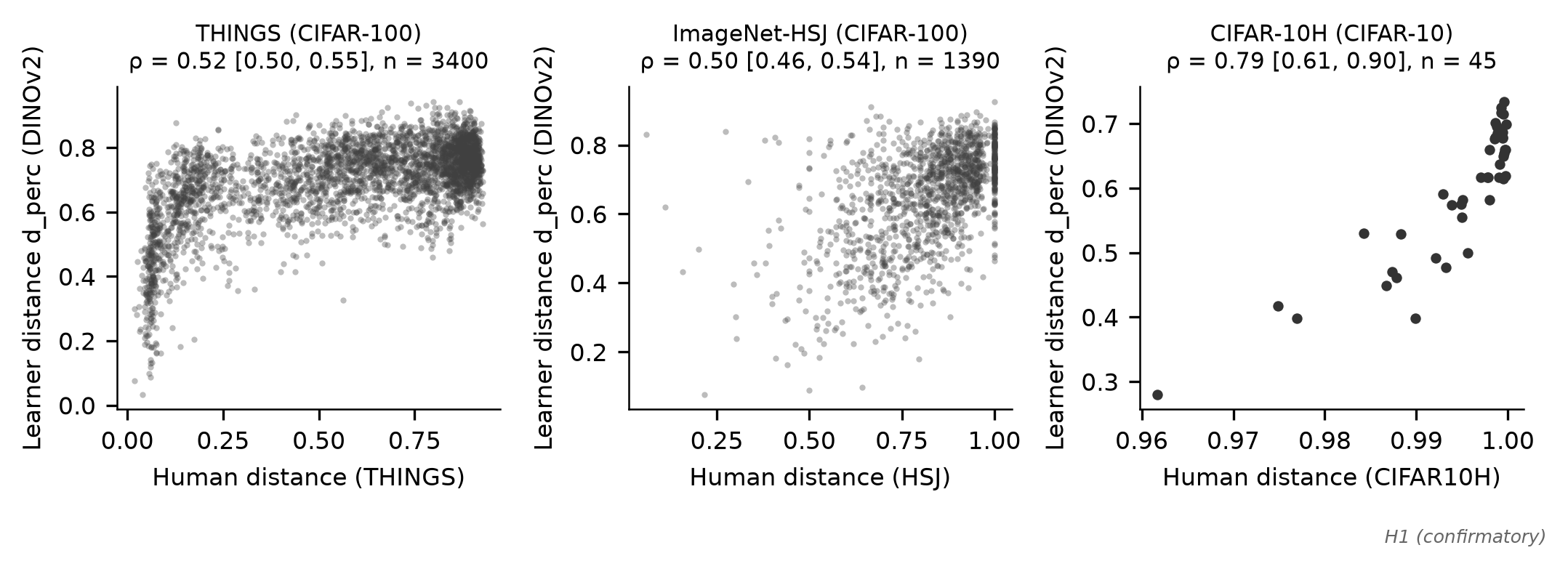}
\caption{Human Distances Versus the Learner's Embedding-Space Distances (H1). \emph{Note.} Each point is a category pair. Brackets give 95\% bootstrap intervals for Spearman's $\rho$.}
\label{fig:h1}
\end{figure}

\paragraph{The encoder's geometry aligns with human measures (H1; supported).} DINOv2 distances correlated with THINGS distances ($\rho = \HOneThingsRho$ [\HOneThingsLo,~\HOneThingsHi], \HOneThingsN{} pairs), ImageNet-HSJ distances ($\rho = \HOneHsjRho$ [\HOneHsjLo,~\HOneHsjHi], \HOneHsjN{} pairs), and CIFAR-10H confusability ($\rho = \HOneCtenRho$ [\HOneCtenLo,~\HOneCtenHi], \HOneCtenN{} pairs); all Mantel tests \parencite{mantel1967} reached the smallest attainable permutation $p$ (\MantelPmin{} with \MantelPerms{} permutations), so the Holm-adjusted $p = \HOneHolmP$ is a bound rather than an estimate (Figure~\ref{fig:h1}). The alignment is partial: a large part of the variance in human distances is not captured.

\paragraph{Alignment across representations.} Table~\ref{tab:h1} gives H1 for every representation. The geometry of every trained encoder correlated positively with the three human datasets ($\rho$ from \VtHoneMin{} to \VtHoneMax), in the support region in every case. So did the geometry of the untrained network, although the correlations were small for THINGS and ImageNet-HSJ: $\rho = \VtHoneRandThings$ [\VbHoneRandThingsLo,~\VbHoneRandThingsHi] for THINGS (Mantel $p = \VbHoneRandThingsP$), \VtHoneRandHsj{} [\VbHoneRandHsjLo,~\VbHoneRandHsjHi] for ImageNet-HSJ ($p = \VbHoneRandHsjP$), and \VtHoneRandCten{} [\VbHoneRandCtenLo,~\VbHoneRandCtenHi] for CIFAR-10H ($p = \VbHoneRandCtenP$). Pixels were positive for THINGS and ImageNet-HSJ ($\rho = \VtHonePixThings$ and \VtHonePixHsj); for CIFAR-10H ($\rho = \VtHonePixCten$, Mantel $p = \VtHonePixCtenP$) the frozen rule ($p < .05$) returned inconclusive. The frozen rule asks only for a positive correlation with Mantel $p < .05$, so it does not distinguish small from large alignments. On THINGS and ImageNet-HSJ, the trained encoders other than MAE reached $\rho$ from \VcHoneTrMin{} to \VcHoneTrMax{} (MAE: \VcHoneMaeThings{} and \VcHoneMaeHsj), against \VtHoneRandThings{} and \VtHoneRandHsj{} for the untrained network, whose upper interval limits (\VbHoneRandThingsHi{} and \VbHoneRandHsjHi) lie below the lowest lower limit among those trained encoders (\VcHoneTrLoMin). The trained encoders were thus far better aligned than the untrained network, even though the binary rule placed all of them in the support region. The frozen global profile for H1, ``holds for all, without evidence of a geometric difference,'' reflects that binary rule and the failure of the geometry calibration at every cutoff (a control shared a geometry group with a trained encoder; Appendix~\ref{app:rules}); it does not mean that the encoders and the untrained network are equally aligned.

\begin{table}[!htbp]
\centering\small
\setlength{\tabcolsep}{3.5pt}
\noindent\adjustbox{max width=\linewidth}{%
\begin{tabular}{llrrr}
\toprule
Encoder & Role & THINGS & ImageNet-HSJ & CIFAR-10H \\
\midrule
DINOv2-small & Descriptive & .52 & .50 & .79 \\
DINOv2-base & Confirmatory & .52 & .55 & .68 \\
CVCL & Descriptive & .68 & .43 & .79 \\
SigLIP & Confirmatory & .64 & .50 & .89 \\
OpenCLIP & Confirmatory & .70 & .52 & .87 \\
Supervised ViT & Confirmatory & .39 & .56 & .66 \\
Supervised ResNet-50 & Confirmatory & .62 & .53 & .91 \\
MAE & Confirmatory & .29 & .23 & .44 \\
\midrule
Random ResNet-50 (s0) & Control & .11 & .11 & .53 \\
Pixels & Control & .15 & .20 & .29$^{\dagger}$ \\
\bottomrule
\end{tabular}}

\caption{Alignment Between Encoder and Human Distances (H1) by Representation}
\label{tab:h1}
\par\smallskip{\raggedright\footnotesize\textit{Note.} Spearman $\rho$ between encoder and human distances (THINGS, \VtHoneNThings{} pairs; ImageNet-HSJ, \VtHoneNHsj; CIFAR-10H, \VtHoneNCten). Support requires $\rho > 0$ and Mantel $p < .05$ (10{,}000 permutations), whatever the size of $\rho$. $^{\dagger}$Not in the support region (Mantel $p = \VtHonePixCtenP$). Full results, including H2, are in Appendix Table~\ref{tab:v2agentsH}.\par}
\end{table}

\paragraph{Human distances do not reliably predict learner difficulty (H2; inconclusive).} A different question is whether human distances, which are independent of the learner's representation, predict which category pairs the learner finds hard. In the dissociation pairs covered by each dataset, with semantic distance in the model, the human coefficient was small for THINGS ($\beta = \HTwoThingsBeta$, $p = \HTwoThingsP$, partial $r = \HTwoThingsR$ [\HTwoThingsRLo,~\HTwoThingsRHi], $n = \HTwoThingsN$) and larger for ImageNet-HSJ ($\beta = \HTwoHsjBeta$, $p = \HTwoHsjP$, Holm $p = \HTwoHsjPHolm$, partial $r = \HTwoHsjR$ [\HTwoHsjRLo,~\HTwoHsjRHi], $n = \HTwoHsjN$). The two results are not symmetric. For THINGS the estimate was small and its interval included zero. For ImageNet-HSJ the effect was in the predicted direction and its interval excluded zero, but it did not survive Holm correction; the plan's decision rule did not anticipate this combination, and we report it as inconclusive without reinterpreting the rule (Appendix Table~\ref{tab:dev}). Both samples are small, which limits power. Across representations, H2 was defined only for THINGS and ImageNet-HSJ, so H2 $\times$ CIFAR-10H is not applicable by design. Because DINOv2-small was inconclusive on THINGS, the multi-encoder comparison for H2 remained descriptive. Its per-encoder regions are unadjusted and test the coefficient (with HC3 standard errors), so they label the DINOv2-small ImageNet-HSJ estimate as support (unadjusted $p = \HTwoHsjP$), whereas the reference analysis above is Holm-adjusted and inconclusive; the two labels answer different questions and both are reported. The same unadjusted rule labels the untrained network as support on ImageNet-HSJ ($p = \VcHtwoRandHsjP$), although the interval of its partial correlation includes zero ($r = \VcHtwoRandHsjR$ [\VcHtwoRandHsjLo,~\VcHtwoRandHsjHi]); with \HTwoHsjN{} pairs such labels are fragile. With \HTwoThingsN{} and \HTwoHsjN{} pairs, the smallest partial correlation detectable with 80\% power at $\alpha = .05$ is \VcMdeThings{} and \VcMdeHsj, respectively. Estimates for every representation, and a descriptive analysis on all mapped pairs, are in Appendix Table~\ref{tab:h2post}.

\begin{figure}[!htbp]
\centering
\includegraphics[width=\linewidth]{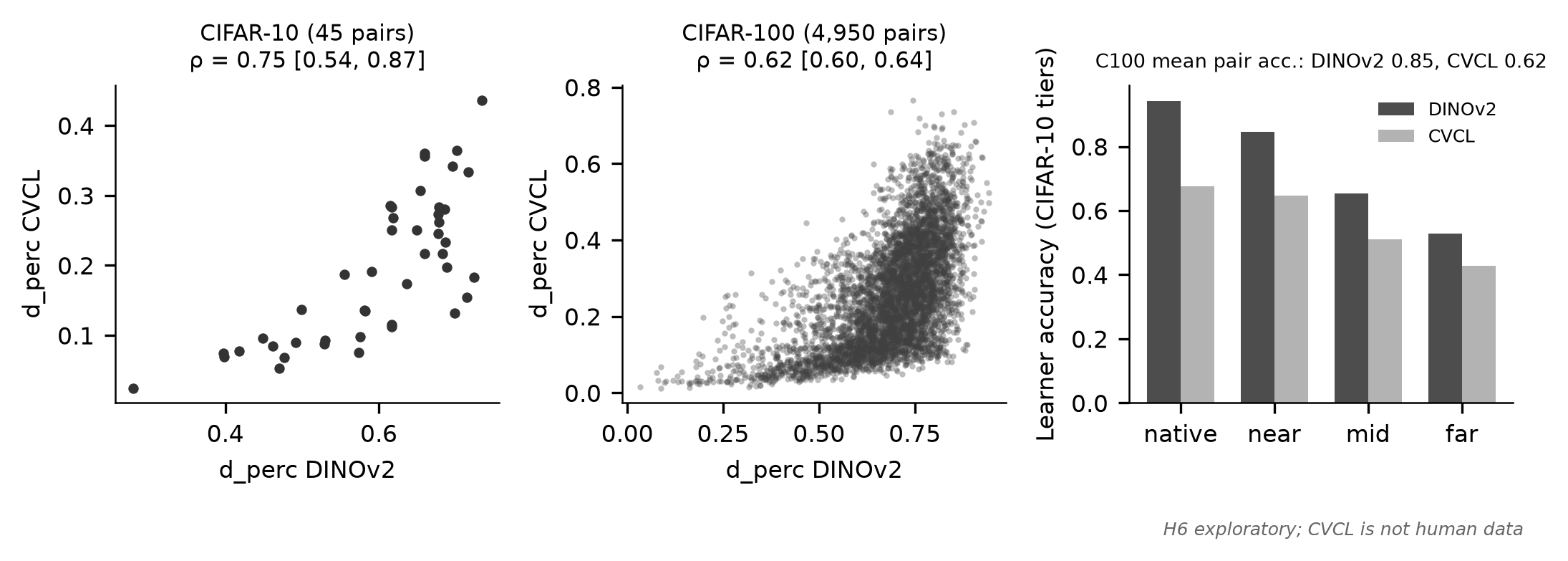}
\caption{DINOv2 Versus a Child-Experience Encoder (CVCL; H6, Exploratory). \emph{Note.} Left and center: pair distances in the two encoders. Right: learner naming accuracy by CIFAR-10 tier with each encoder.}
\label{fig:h6}
\end{figure}

\paragraph{A child-experience encoder reproduces the tiers but not the pair-level effect (H6; exploratory).} To check whether the learner results depend on DINOv2, we repeated the analyses with the vision encoder of CVCL, trained on one child's head-camera recordings \parencite{vong2024}; CVCL is not a source of human data. Images were resized directly to 224 $\times$ 224 pixels (bicubic, no crop), converted to tensors, and normalized with ImageNet means and standard deviations, following the preprocessing in the released CVCL code; the 512-dimensional outputs were L2-normalized. Unlike the DINOv2-small pipeline, which keeps the central 224 of 256 pixels per side of the upsampled image, this preprocessing keeps the full frame, so the field of view also differs between the two encoders. Its distances correlated with DINOv2 ($\rho = \HSixRhoTen$ in CIFAR-10, \HSixRhoHundred{} in CIFAR-100) and with human distances (e.g., THINGS \HSixThingsHundred, HSJ \HSixHsjHundred{} in CIFAR-100). With CVCL, the tier ordering reappeared in CIFAR-10 (\CvclNative\%, \CvclNear\%, \CvclMid\%, \CvclFar\%; $\rho = \CvclTierRho$ across words), but in CIFAR-100 its pair distances did not predict naming accuracy ($\beta = \HSixB$ [\HSixLo,~\HSixHi], $p = \HSixP$; the multi-encoder dissociation analysis D for CVCL in Table~\ref{tab:agents} is a separate computation, reported as a partial correlation, $r = \VtDCvclR$), which is inconsistent with the pre-specified criterion (Benjamini--Hochberg correction over the \HSixBHTests{} H6 tests; Figure~\ref{fig:h6}). Baseline naming accuracy in CIFAR-100 was also lower with CVCL (\HSixBaseCvcl{} vs.\ \HSixBaseDino{} with DINOv2), which limits the comparison; the CIFAR-10 tier ordering is descriptive and was not tested inferentially.

\paragraph{Age of acquisition (H5; exploratory).} Category coherence in DINOv2 was not related to the age of acquisition of the category names in the Glasgow norms \parencite{scott2019} in CIFAR-100 (partial $\rho$ controlling log word frequency $= \HFiveHundredRho$, $p = \HFiveHundredP$, $n = \HFiveHundredN$); in CIFAR-10 the partial correlation was \HFiveTenRho{} (BH $p = \HFiveTenPBH$), but with only ten categories. We do not draw conclusions from H5.

\section{Adults and Agents on the Same Images (Exploratory)}
\label{sec:matched}

\paragraph{Units of analysis.} The comparisons in this article combine three different units. Adults contribute judgments per session; each agent contributes one decision per image and word; and the CIFAR distance analyses use pairs of classes. Counts and coefficients in different tables are therefore not on a common scale.

This comparison was exploratory. It was specified after the adult outcomes and the motivating image pairs were known, and it was run before the multi-encoder learner analyses were frozen (timestamps in Appendix~\ref{app:deviations}); its protocol and outputs were committed together and were not frozen separately. Each agent was a centroid learner over one encoder, fitted to the same teaching images that adults saw in \textcite{lewis2018} and deciding on the same 24 test images with a support-dependent acceptance rule. Adults contribute selections in \VtXPerCell{} sessions per word--image pairing; each agent makes one deterministic decision per pairing.

\paragraph{Decision rules.} The centroid score of an image is its cosine similarity to the normalized centroid of the teaching images ($1 - d$). With three teaching images $s_1, s_2, s_3$ and a candidate $x$, the learner was refitted on each set of three images that leaves out one teaching image and scored the image left out; $x$ was accepted if at least one teaching image scored at or below $x$ (tolerance $10^{-6}$), that is, if $x$ was at least as typical as the least typical teaching image. This support-dependent rule is the one written in the protocol; it uses no human response or category label, can accept any number of the 24 images, and lets a dispersed teaching set broaden acceptance by itself. With one teaching image it is undefined, so one-example trials were not used. Separately, for each encoder we computed the largest gain attainable with one \emph{common threshold}: the 12 within-domain superordinate candidates (four per domain) were scored after broad and after same-subtype teaching, a single threshold on the centroid score was applied to both conditions and all three domains, and the gain is the maximum, over all thresholds, of the fraction accepted after broad teaching minus the fraction after same-subtype teaching, with no constraint on other-domain acceptance. It is a diagnostic upper bound, not a chosen operating point. The adult benchmark is the same contrast, \VtXBroadPct\% minus \VcXSameSubPct\% (same-subtype teaching). With 12 candidates, rates move in steps of 1/12 (\VcXStep{} points): the largest encoder gain, \VtXJumpMax{} points (8 of 12), is one candidate below 9 of 12 (\VcXNextStep{} points), the step closest to the adult value. The same bound with the two other deterministic learners of the protocol ranged from \VcXJumpKcMin{} to \VcXJumpKcMax{} points (two centroids) and from \VcXJumpKnnMin{} to \VcXJumpKnnMax{} points (weighted 3-nearest-neighbor exemplar learner) across the trained encoders. A three-word forced-choice version, in which each image receives one of three taught words, is in the released outputs. We did not try calibrated thresholds, normalized scores, or threshold-free measures.

Table~\ref{tab:v2matched} shows extension by teaching condition. After three examples of the same basic type, adults selected other members of the broader category on \VtXSameBasicPct\% of judgments; after three examples from different basic types of the domain, on \VtXBroadPct\%; images from other domains remained at \VtXOutBroadPct\% after broad teaching. Supervised ViT and DINOv2-base each accepted \VtXViTIn{} of \VtXInN{} within-domain candidates after broad teaching, but ViT accepted \VtXViTOut{} of the \VtXOutN{} other-domain pairings and DINOv2-base accepted \VtXDinoBOut; under the same rule, the two encoders operate at very different false-alarm rates. Across the eight trained encoders, the largest narrow-to-broad gain attainable with a common threshold on the centroid score ranged from \VtXJumpMin{} to \VtXJumpMax{} percentage points, against \VtXJumpAdult{} points for adults.

\begin{figure}[!htbp]
\centering
\includegraphics[width=0.45\linewidth]{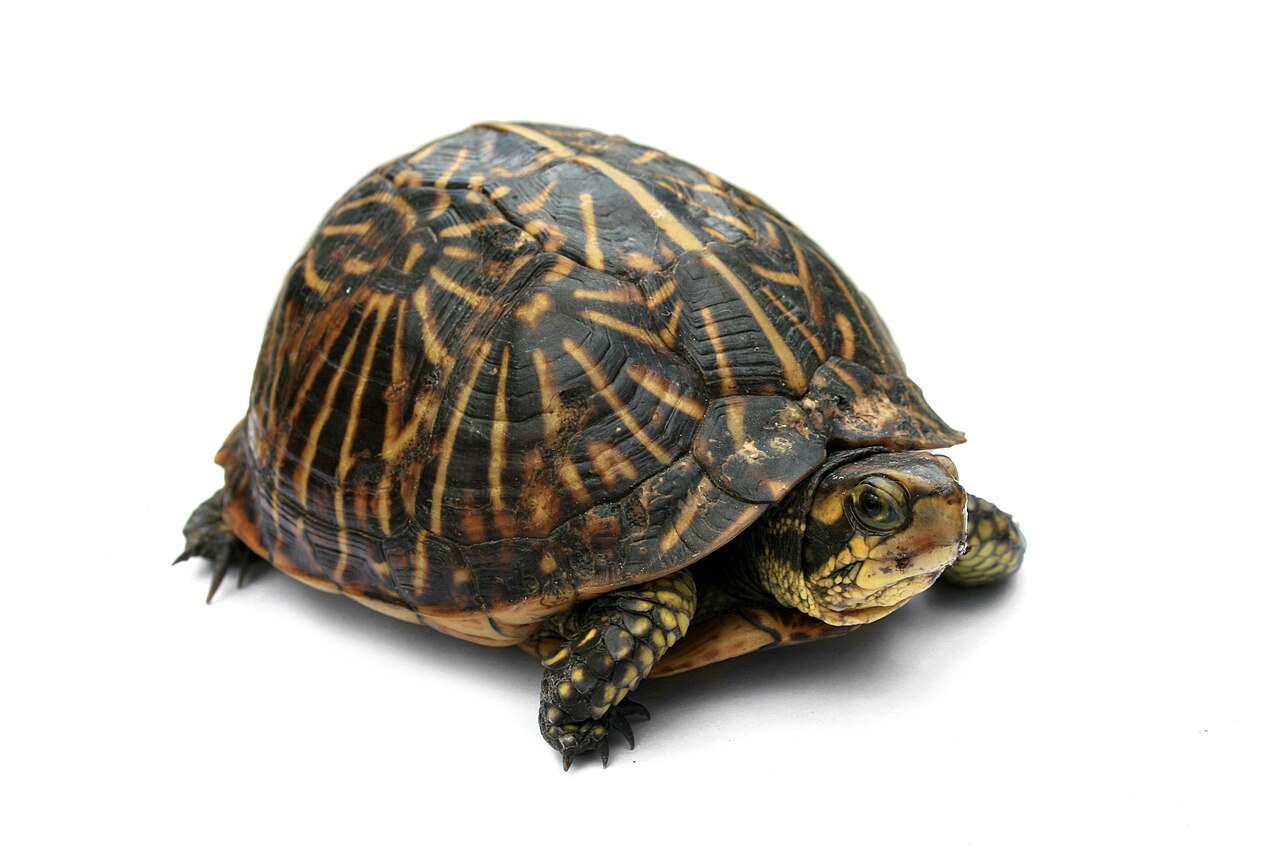}
\caption{Illustrative Photograph of a Turtle}
\label{fig:v2turtle}
\par\smallskip{\raggedright\footnotesize\textit{Note.} Illustrative image only (a Florida box turtle, \textit{Terrapene carolina bauri}). It is \emph{not} the stimulus \texttt{c3\_sup2} shown to participants, which is not reproduced here; it only conveys the kind of image (a whole turtle, side view, plain background). Photo: Jonathan Zander, CC BY-SA 3.0, via Wikimedia Commons, \url{https://commons.wikimedia.org/wiki/File:Florida_Box_Turtle_Digon3.jpg}. Resized; otherwise unmodified. The photograph is licensed under CC BY-SA 3.0 (\url{https://creativecommons.org/licenses/by-sa/3.0/}); this license applies to the photograph only, and any adaptation of it must be shared under the same license.\par}
\end{figure}

The central example is a single photograph (\texttt{c3\_sup2}, a turtle). In the exploratory matched-stimulus comparison, adults selected the same turtle photograph in \VtTurtleAnimalsK{} of \VtTurtleSessions{} sessions after broad animal teaching and in \VtTurtleVehiclesK{} of \VtTurtleSessions{} after broad vehicle teaching. DINOv2-base accepted the turtle in both contexts, while supervised ViT rejected it in both. The three adult counts come from the same \VtTurtleSessions{} sessions, because each broad-teaching session had one trial per domain. Paired within session, \VtTurtleOnlyAnimals{} sessions selected the turtle only under the animal word, \VtTurtleOnlyVehicles{} only under the vehicle word, \VtTurtleBoth{} under both, and \VtTurtleNeither{} under neither; under the vegetable word it was selected in \VtTurtleVegK{} of \VtTurtleSessions{} sessions (Table~\ref{tab:v2turtle}; Figure~\ref{fig:v2turtle} shows an illustrative turtle photograph, not the stimulus). The two agents thus departed from the adult pattern in opposite directions.

Table~\ref{tab:v2images} lists the decision of each trained encoder on each of the 12 within-domain candidates after broad teaching. It describes which images received the novel word in this task. For example, supervised ResNet-50 accepted \VtXEthreeAnimals{} of 4 animal and \VtXEthreeVeg{} of 4 vegetable candidates and \VtXEthreeVeh{} of 4 vehicle candidates, whereas adults selected even the least chosen vehicle candidate in \VtXRailcarK{} of \VtXPerCell{} sessions; this is a statement about which images received the word, not about whether the network recognizes vehicles.

\begin{table}[!htbp]
\centering\small
\setlength{\tabcolsep}{4pt}
\noindent\adjustbox{max width=\linewidth}{%
\begin{tabular}{lcccccc}
\toprule
 & \multicolumn{3}{c}{Within-domain superordinates} & \multicolumn{3}{c}{Other-domain images} \\ \cmidrule(lr){2-4}\cmidrule(lr){5-7}
Learner & Same subtype & Same basic type & Broad & Same subtype & Same basic type & Broad \\
\midrule
Adults (selections / judgments) & 76/7200 & 132/7200 & 5579/7200 & 177/28800 & 83/28800 & 97/28800 \\
\midrule
DINOv2-small & 0/12 & 3/12 & 7/12 & 0/48 & 2/48 & 6/48 \\
DINOv2-base & 0/12 & 6/12 & 10/12 & 0/48 & 12/48 & 20/48 \\
CVCL & 0/12 & 6/12 & 11/12 & 0/48 & 5/48 & 14/48 \\
SigLIP & 0/12 & 0/12 & 5/12 & 0/48 & 0/48 & 8/48 \\
OpenCLIP & 0/12 & 0/12 & 6/12 & 0/48 & 0/48 & 1/48 \\
Supervised ViT & 0/12 & 3/12 & 10/12 & 0/48 & 0/48 & 0/48 \\
Supervised ResNet-50 & 0/12 & 4/12 & 8/12 & 0/48 & 2/48 & 8/48 \\
MAE & 0/12 & 9/12 & 8/12 & 0/48 & 22/48 & 23/48 \\
\midrule
Random ResNet-50 (s0) & 1/12 & 9/12 & 9/12 & 4/48 & 30/48 & 38/48 \\
Pixels & 4/12 & 9/12 & 9/12 & 14/48 & 35/48 & 20/48 \\
\midrule
MAE (CLS) & 0/12 & 2/12 & 11/12 & 0/48 & 0/48 & 16/48 \\
Random ResNet-50 (s1) & 2/12 & 7/12 & 8/12 & 8/48 & 31/48 & 35/48 \\
Random ResNet-50 (s2) & 2/12 & 7/12 & 6/12 & 10/48 & 25/48 & 30/48 \\
\bottomrule
\end{tabular}}

\caption{Exploratory: Extension by Teaching Condition, Adults and Agents on the Same Images}
\label{tab:v2matched}
\par\smallskip{\raggedright\footnotesize\textit{Note.} Exploratory; specified after the adult outcomes were known and run before the agent freeze. Cells give selections or acceptances over pairings: within-domain superordinates (\VtXInDenom{} adult judgments, 12 agent decisions per condition) and other-domain images (\VtXOutDenom{} adult judgments, 48 agent decisions). Adult judgments are repeated within person; agent decisions are deterministic. Centroid learner with a support-dependent acceptance rule.\par}
\end{table}

\begin{table}[!htbp]
\centering\small
\setlength{\tabcolsep}{4pt}
\noindent\adjustbox{max width=\linewidth}{%
\begin{tabular}{lllccccccccc}
\toprule
Taught domain & Image ID & Visual description & Adults ($k$/600) & DINOv2-s & DINOv2-b & CVCL & SigLIP & OpenCLIP & ViT & RN50 & MAE \\
\midrule
Animals & \texttt{c3\_sup1} & rabbit & 490 & A & A & A & A & A & A & A & A \\
 & \texttt{c3\_sup2} & turtle & 431 & A & A & A & R & R & R & A & A \\
 & \texttt{c3\_sup3} & horse & 496 & R & A & A & A & R & A & A & A \\
 & \texttt{c3\_sup4} & cat & 485 & R & A & A & A & R & A & A & A \\
\midrule
Vegetables & \texttt{c1\_sup1} & celery & 494 & A & A & A & R & A & A & A & A \\
 & \texttt{c1\_sup2} & potato & 418 & R & R & A & R & R & A & A & A \\
 & \texttt{c1\_sup3} & radish & 502 & A & A & A & R & R & A & A & A \\
 & \texttt{c1\_sup4} & broccoli & 480 & R & R & R & A & A & A & A & R \\
\midrule
Vehicles & \texttt{c2\_sup1} & dump truck & 499 & A & A & A & A & A & A & R & R \\
 & \texttt{c2\_sup2} & red railway car & 374 & A & A & A & R & A & R & R & R \\
 & \texttt{c2\_sup3} & blue car & 452 & A & A & A & R & R & A & R & R \\
 & \texttt{c2\_sup4} & gray car & 458 & R & A & A & R & A & A & R & A \\
\midrule
Accepted &  &  & 5579/7200 & 7/12 & 10/12 & 11/12 & 5/12 & 6/12 & 10/12 & 8/12 & 8/12 \\
\bottomrule
\end{tabular}}

\caption{Exploratory: Decisions on the 12 Within-Domain Candidates After Broad Teaching}
\label{tab:v2images}
\par\smallskip{\raggedright\footnotesize\textit{Note.} Exploratory. A = accepted; R = rejected (centroid learner). Image IDs are those of the original stimulus set \parencite{lewis2018}; descriptions are visual descriptions of the photographs, not category labels used in the task. DINOv2-s, DINOv2-b, ViT, and RN50: DINOv2-small, DINOv2-base, supervised ViT, and supervised ResNet-50. Adults: selections in \VtXPerCell{} sessions per image. The table describes which images received the novel word in this task.\par}
\end{table}

\begin{table}[!htbp]
\centering\small
\setlength{\tabcolsep}{4pt}
\noindent{%
\begin{tabular}{lccccccccc}
\toprule
Context & Adults ($k$/600) & DINOv2-s & DINOv2-b & CVCL & SigLIP & OpenCLIP & ViT & RN50 & MAE \\
\midrule
Animals & 431 & A & A & A & R & R & R & A & A \\
Vehicles & 1 & R & A & R & R & R & R & R & A \\
Vegetables & 3 & R & R & A & R & R & R & A & R \\
\bottomrule
\end{tabular}}

\caption{Exploratory: The Turtle Photograph (\texttt{c3\_sup2}) Under the Three Broad-Teaching Words}
\label{tab:v2turtle}
\par\smallskip{\raggedright\footnotesize\textit{Note.} Exploratory. Adults: selections in the same \VtTurtleSessions{} sessions under each word. A = accepted; R = rejected. Abbreviations as in Table~\ref{tab:v2images}. DINOv2-base accepted \VtXDinoBOut{} of \VtXOutN{} other-domain pairings after broad teaching, so its acceptance of the turtle under the vehicle word partly reflects a permissive operating point of the rule for this encoder.\par}
\end{table}

\section{Discussion}

We asked how much visual similarity explains when two objects share a name, and what the teaching examples add. In adults, distance in a self-supervised visual space was associated with the decision to extend a novel word across specifications, and, in an exploratory post hoc test, it added to taxonomic level and teaching condition within the taught domain, although its interval included zero once the slope could vary across images (post hoc). In artificial learners operating over the same space, words joining more distant categories were harder to learn, and taxonomic relatedness did not predict accuracy beyond perceptual distance in this sample (only \NWithin{} of 100 pairs fell within a superclass); because the learner has no access to semantic information, this is a sufficiency result. The encoder's geometry aligned partially with three independent human measures of similarity and confusability. Taken together, these results suggest that distances in these spaces describe part of the organization of generalization in both systems. For the learners this is close to mechanical, since they operate on the same distances, and the direct test of whether human distances predict learner difficulty (H2) was inconclusive, so the link between the two systems remains indirect.

Across representations, the picture is consistent but limited. Within the taught domain, every trained encoder, and also a randomly initialized network, gave a negative distance slope, so the association cannot be attributed only to what the encoders learned (with by-image random slopes, post hoc, the slopes varied widely across images and the interval included zero for \VcImgTrInclList); the frozen primary test of whether presentation order changes that slope was inconclusive. In the learners, the distance effect appeared for most but not all trained encoders, and the binary H1 rule did not separate the trained encoders from the untrained network, although the correlations of the trained encoders other than MAE with THINGS and ImageNet-HSJ were several times larger. We did not test which properties of an untrained convolutional network, absent from raw pixels, produce its negative slope. In the exploratory comparison on the same images, learners with equal within-domain acceptance extended the word to different images, and no common threshold on a trained encoder reproduced the adults' narrow-to-broad gain.

The limits are as informative as the associations. First, in adults the composition of the examples changed extension dramatically at comparable distances: three superordinate examples licensed broad extension, whereas three subordinate examples suppressed it, and, in an exploratory analysis, the slope of distance itself depended on the teaching condition. This is the kind of evidence-sensitive inference that Bayesian accounts emphasize \parencite{xu2007} and that \textcite{lewis2018} showed to be sensitive to trial order; the distance-to-centroid measure we used does not, by itself, account for these differences between conditions, and in post hoc checks neither did the mean or nearest distance to the individual examples (Appendix~\ref{app:posthoc}). These checks do not exhaust readouts of the same geometry: models that scale generalization by the dispersion of the examples \parencite{tenenbaum2001} remain untested. In our exploratory post hoc models, terms involving experiment-level trial order improved fit in observation-level tests that ignore the clustering of decisions within experiments, and within superordinate items both condition and distance improved fit given the other (distance no longer did once its slope could vary across images); because order is confounded with other experiment-level differences, we do not interpret these terms further. Second, representational alignment did not translate into conclusive evidence that human distances predict learner difficulty: the encoder's geometry aligned partially with human measures (H1), but human-derived distances did not reliably predict which words the learner found difficult (H2, inconclusive; one dataset showed an effect in the predicted direction that did not survive correction). Correlation between two representational spaces does not guarantee that both predict learning in the same way. Third, the pair-level learner effect did not reappear with an encoder trained on child experience, so it may depend on properties of the specific encoder. And because THINGS and ImageNet-HSJ mix perceptual and conceptual information, the alignment we observe cannot be attributed to perceptual similarity alone.

We therefore claim similarity in the organization of generalization, not a shared mechanism. The adult and the learner performed different tasks, measured on different scales (Figure~\ref{fig:x1}), and the learner has no access to the pragmatic and sampling considerations that adults bring to the task. The value of the artificial system is that it isolates one factor, similarity in a known space, and shows what that factor alone can and cannot produce.

\subsection{Limitations}

All human data are secondary. The experiments were designed by other teams for other questions, with different stimuli, tasks, and participants, and we could not alter their designs. The adult analysis is correlational: distance and taxonomic level are related by construction in the stimulus sets, and although models control for level and condition, distance in any one encoder is one of many possible similarity measures, and ten representations do not exhaust them. The descriptive contrast between conditions and the interaction model are exploratory. The H2 analyses had small samples (\HTwoThingsN{} and \HTwoHsjN{} pairs). All learner results rest on frozen encoders and CIFAR images, two-category words, and shallow learners; only one final-layer global embedding per encoder was used, read out as the cosine distance to the normalized centroid. Intermediate layers and patch tokens, other metrics (e.g., Mahalanobis or whitened Euclidean distance), exemplar models of categorization such as the generalized context model \parencite{nosofsky1986}, and a randomly initialized vision transformer as a second untrained control were not examined; the post hoc exemplar distances of Appendix~\ref{app:posthoc} are only a first check. The planned human mixed models include random intercepts by image (24 stimulus slots) but no by-image random slopes for distance; in a post hoc check with by-image slopes, the image slopes varied widely and the interval of the average slope included zero for \VcImgTrInclList{} (Appendix~\ref{app:posthoc}). For comparability across encoders, the stage-1 models of the primary test used the structure that converged for every encoder, which had by-person random intercepts only. The dissociation pairs were selected with DINOv2-small, and the WordNet sense correction was checked only with DINOv2-small. Preprocessing pipelines differ in field of view on CIFAR images (Appendix~\ref{app:checkpoints}). The CIFAR-10 tier intervals are episode-level and do not separate pair-level variability. Taxonomic relatedness was measured only with WordNet Wu--Palmer similarity, with few within-superclass pairs. The two post hoc analyses of trial order and superordinate items were specified before being run but after the data had been analyzed, and trial order varied between experiments, together with other design features. The previous multi-agent consensus experiments of this project are reported in \textcite{vera2026lexical} and are not part of this article.

The frozen primary test of order moderation was inconclusive, so it supports neither moderation nor its absence. The distance association in adults also appeared with a randomly initialized network, and the binary H1 rule did not separate trained encoders from the controls, despite large differences in the size of the correlations. The \VtDallN-pair sensitivity analysis uses pairs that share classes. The comparison on the same images is exploratory, was specified after the adult outcomes were known, and rests on \VtXInN{} and \VtXOutN{} deterministic decisions per agent and teaching condition, so a single decision moves a within-domain rate by one twelfth; it uses one stimulus set with three domains. The optional variants (MAE with CLS pooling, random-network seeds 1 and 2) were not run in the CIFAR analyses; in adults they entered only the sensitivity analyses of the primary test, and they are reported in the exploratory comparison on the same images (Table~\ref{tab:v2matched}).

\section{Conclusions and Future Work}

\subsection{Conclusions}

In the planned within-domain models, greater cosine distance from the test image to the centroid of the teaching examples was associated with less word extension for all eight trained encoders and for the randomly initialized network. In post hoc models that allowed the distance slope to vary across images, \VcImgTrSameSignWord{} of the eight trained encoders retained a negative estimate, and \VcImgTrExclWord{} had 95\% intervals entirely below zero (Appendix~\ref{app:posthoc}). In an exploratory comparison, presenting adults and agents with the same images illustrated two features of generalization: \textbf{when a learner broadens a word’s scope and where that extension stops}. After three examples of the same subtype, adults selected other members of the broader category on only \VcXSameSubPct\% of judgments. After three examples drawn from different basic types within one domain, that rate rose to \VtXBroadPct\%, while selection outside the domain remained at \VtXOutBroadPct\%. After broad teaching, the agents using supervised ViT and DINOv2-base each accepted \VtXViTIn{} of \VtXInN{} within-domain images, agreeing on \VdAgreeWord{} and differing on \VdDifferWord. Outside the taught domain, supervised ViT accepted none of the \VtXOutN{} word--image pairings, whereas DINOv2-base accepted \VtXDinoBOut. The adult pattern is consistent with an inference about the intended scope of a novel word. Whether adults arrive at that inference through the same processes as any of these agents remains open.

The internally frozen primary analysis did not establish whether teaching order changes the slope of visual distance: for DINOv2-small, the experiment-level slope difference was \VtOrdDelta{} (95\% CI \VtOrdLo{} to \VtOrdHi), and the frozen decision rule returned an inconclusive global result at every combination of effect margin and geometry cutoff. In the CIFAR agents, greater visual distance between the two classes predicted lower naming accuracy for \VtDSupportWord{} of the eight trained encoders (partial correlations from \VtDrMin{} to \VtDrMax{} on the \VtDN{} original class pairs), but not for CVCL ($r = \VtDCvclR$, 95\% CI \VtDCvclLo{} to \VtDCvclHi) or MAE ($r = \VtDMaeR$, 95\% CI \VtDMaeLo{} to \VtDMaeHi); these two moved into the support region only in the descriptive analysis of all \VtDallN{} class pairs, which share classes and are not independent. The geometry of every trained encoder correlated positively with human similarity judgments on THINGS, ImageNet-HSJ, and CIFAR-10H (Spearman $\rho$ from \VtHoneMin{} to \VtHoneMax), but so did the geometry of a randomly initialized network ($\rho = \VtHoneRandThings$, \VtHoneRandHsj, and \VtHoneRandCten), and the pixel baseline was positive on THINGS and ImageNet-HSJ ($\rho = \VtHonePixThings$ and \VtHonePixHsj) and inconclusive on CIFAR-10H ($\rho = \VtHonePixCten$, Mantel $p = \VtHonePixCtenP$); the binary support rule did not separate the encoders from the controls, although on THINGS and ImageNet-HSJ the trained encoders other than MAE correlated far more strongly with human judgments ($\rho$ from \VcHoneTrMin{} to \VcHoneTrMax) than the untrained network. In the exploratory matched-image comparison, which was specified after the adult outcomes were known, no common threshold on a trained encoder's centroid score reproduced the adult narrow-to-broad gain: the largest attainable gain ranged from \VtXJumpMin{} to \VtXJumpMax{} percentage points across the eight trained encoders, against \VtXJumpAdult{} points for adults (same-subtype to broad teaching); the largest value is one candidate in twelve below the next attainable step. Post hoc, distances to the individual teaching examples did not account for the effect of composition in adults either.

\subsection{Future Work}

A next step is to train agents to use relations among categories to infer how broadly a novel word applies. A bus, a motorcycle, and a tractor provide evidence that a word may denote \emph{vehicle}. Three buses also belong to that category, but provide weaker evidence that the speaker intends the broader meaning. This distinction can be tested by holding the visual encoder and training images fixed while varying whether the learner receives information about relations among subtypes, basic types, and broader categories. At test, both learners would encounter novel words and previously unseen category members. The critical case is one in which visual resemblance and category membership favor different choices. A learner that uses category relations should extend the word to a visually dissimilar member after broad teaching, reject a visually similar outsider, and preserve a narrower extension after same-type teaching. Transfer of this pattern to new cases would provide evidence that learned relations help the agent interpret what a word refers to. A direct test would ask whether training on category relations improves transfer to novel items beyond what a learner can achieve by adapting to the visual dispersion of the same teaching examples.

Other directions follow from the limits of the present readout: intermediate layers and patch tokens, other metrics (Mahalanobis or whitened Euclidean distances), exemplar models of categorization such as the generalized context model \parencite{nosofsky1986}, dispersion-sensitive accounts of generalization \parencite{tenenbaum2001}, and a randomly initialized vision transformer as a second untrained control.

\section{Ethics Statement}

No new data were collected for this article. The human analyses used only publicly available, anonymized data released by the authors of published studies \parencite{lewis2018,hebart2023,roads2021,peterson2019,scott2019}. Ethical approval and participant consent were handled by the original studies, and no attempt was made to re-identify participants.

\section{Code and Data Availability}

Code, analysis plans, logs, the deviation log, and figure-generation scripts for the artificial-learner experiments and for the human-data analyses (folder \texttt{human\_data\_analyses/}) are available at \url{https://github.com/patriciomvera/lexical-consensus}. The hashes of the analysis plans are listed in Appendix~\ref{app:deviations}. Derived datasets released with this article are available on Hugging Face: \url{https://huggingface.co/datasets/MrPatoVera/lexical-consensus-data} (CC BY 4.0) and\newline \url{https://huggingface.co/datasets/MrPatoVera/lexical-consensus-data-cifar10h} (CC BY-NC-SA 4.0). Human datasets are publicly available from their original authors \parencite{lewis2018,hebart2023,roads2021,peterson2019,scott2019}; files that are not redistributed are downloaded by a script from their official sources. Licenses, as stated by the original sources: the data of \textcite{lewis2018} are released on the authors' OSF project under CC BY 4.0 (the GitHub copy carries no license file); we do not redistribute these data or any per-participant or per-trial derivative of them, and the scripts download them from the source. THINGS odd-one-out data are released under CC BY 4.0 (with exceptions for some images, which we do not use or redistribute); ImageNet-HSJ (psiz-datasets) under CC BY 4.0; the Glasgow Norms in an article published under CC BY 4.0; CIFAR-10H under CC BY-NC-SA 4.0; and the wordfreq data under CC BY-SA 4.0. Derived files that include CIFAR-10H aggregates or wordfreq frequencies are marked in the repository as share-alike (and, for CIFAR-10H, non-commercial); CIFAR images are not redistributed. CIFAR-10/100 \parencite{krizhevsky2009}, DINOv2 \parencite{oquab2023}, CVCL \parencite{vong2024}, and WordNet \parencite{miller1995} are publicly available, as are the other encoders at the revisions listed in Appendix~\ref{app:checkpoints}.

The multi-encoder analyses, freeze manifests, deviation logs, item-level outputs, and exhibit scripts are in the folder \texttt{visual\_similarity\_v2/} of the same repository, cited here at commit \texttt{a945ed2}: \url{https://github.com/patriciomvera/lexical-consensus/tree/a945ed2cbdd9365fe346b262763f0262eaf686ea/visual_similarity_v2}. The CIFAR embedding caches are not redistributed; their manifest is. All tables, figures, and numbers of the multi-encoder analyses are generated by scripts from those files. The post hoc analyses of Appendix~\ref{app:posthoc} are not yet part of that commit; their scripts and outputs will be added to the repository.

{\emergencystretch=2em\printbibliography}

\appendix
\section{Analysis Plans, Deviations, and Timestamps}
\label{app:deviations}

\paragraph{Plans and hashes.} The analysis plan for the human-data analyses with DINOv2-small was fixed before the confirmatory analyses, after a check of dataset coverage, and was not deposited in an external registry (sha256 \texttt{\PlanHash\ldots}). The hash identifies the version of the plan that was used, but it does not certify when the plan was written; we therefore do not describe it as a preregistration. A copy that differs only in the replacement of a personal name, prepared for an earlier anonymized review, has sha256 \texttt{\PlanHashAnon\ldots}. The multi-encoder plans were frozen by checksum manifests released in the repository. The learner analyses were frozen on \VtFreezeDate{} at \VtFreezeTime{} (UTC$-$3); the exploratory comparison on the same images was run on \VtXtDate{} at \VtXtTime{} (UTC$-$3), before that freeze. Table~\ref{tab:dev} lists the main deviations in the DINOv2-small human-data analyses and Table~\ref{tab:devtwo} those in the multi-encoder analyses; the released deviation logs document every step. Hypotheses are numbered as in the first plan: H1 (alignment with human distances), H2 (human distances and learner difficulty), H3 (a descriptive comparison of the CIFAR-10 tiers with human distances, not reported here), H4 (adult word extension), H5 (age of acquisition, exploratory), and H6 (the CVCL encoder, exploratory). Appendix~\ref{app:rules} defines the decision rules and Appendix~\ref{app:posthoc} reports the analyses added after review.

\begin{table}[!htbp]
\centering\footnotesize
\begin{tabular}{p{0.28\linewidth}p{0.64\linewidth}}
\toprule
Deviation & Description and consequence \\
\midrule
Random effects by person (H4) & The plan specified random effects by participant identifier (\texttt{subid}). Our initial implementation grouped by experiment $\times$ identifier, on the mistaken assumption that identifiers were reused across experiments for different people; they denote persons across experiments (\HFourPersons{} people, \HFourSessions{} sessions). All H4 models were re-estimated with random effects by person, as the plan specified, and these are reported. Conclusions unchanged. \\
First session (H4 sensitivity) & Defined by the recorded start time of each session rather than by experiment number; the two criteria differ for \FirstSessDiff{} person. \\
Teaching-condition labels (exploratory) & Equivalent condition labels spelled differently across experiments were unified in the descriptive analyses after an error was found; H4 models were not affected. \\
Coverage & THINGS: \CovThingsPairs{} category pairs and \HTwoThingsN{} sampled pairs, versus \CovThingsPairsPlan{} and \CovThingsNPlan{} planned, due to category mapping. Glasgow norms: \HFiveHundredN{} words versus \CovGlasgowPlan{} planned. \\
H2 decision rule & The ImageNet-HSJ result (uncorrected $p = \HTwoHsjP$, Holm $p = \HTwoHsjPHolm$, interval excluding zero) was not covered by any of the plan's decision rules; reported as inconclusive. \\
Analyses not run & Wordbank analyses (data available only through an authenticated service) and overextension analyses (unclear license). \\
H6 correction family & Benjamini--Hochberg correction over \HSixBHTests{} tests; the plan did not specify how the family was composed. \\
Post hoc H4 analysis & Incremental contribution of distance within the taught domain (the distance $\times$ condition interaction was planned as exploratory and is not a deviation); exploratory, outside the confirmatory family. \\
Post hoc analyses after external review & Moderation by trial order and a superordinate-only model, specified after an automated external review and after the main results were known; exploratory, outside the confirmatory family. \\
\bottomrule
\end{tabular}
\caption{Main Deviations From the Analysis Plan for the Human Data (DINOv2-small Analyses)}
\label{tab:dev}
\end{table}

\begin{table}[!htbp]
\centering\footnotesize
\begin{tabular}{p{0.28\linewidth}p{0.64\linewidth}}
\toprule
Deviation & Description and consequence \\
\midrule
Pipeline control 1 tolerance & Pipeline control 1 checks that the agent code with DINOv2-small reproduces the earlier CIFAR-10 tier results exactly; pipeline control 2 checks that re-encoding CIFAR-100 with DINOv2-small on CPU reproduces the earlier distance matrix. The rule text required a maximum margin difference $\le 10^{-6}$; the frozen code accepted $\le 10^{-6} + 10^{-12}$ (line \VtCtrlOneLine{} of the agent simulation script), a margin agreed before the run to absorb sixth-decimal rounding. The observed value was $\VtCtrlOneObs$; the control passed under the frozen implementation. Pipeline control 2 passed (maximum difference $\VtCtrlTwoMax$ over \VtCtrlTwoPairs{} pairs). \\
Checksum of a historical patch & The archived file \texttt{DIFF\_agentes\_propuesta.patch} (a proposal that was never applied) is published with sha256 \texttt{\VtPatchPub}; the agent freeze manifest (line \VtPatchFrzLine) expects \texttt{\VtPatchFrz}. The two differ only in a \VtPatchHdrLines-line header marking the file as obsolete and not applied; the difference predates the consolidation of the archive. No analysis reads this file. \\
Optional sensitivities not run & CIFAR analyses with MAE CLS pooling and with random ResNet-50 seeds 1 and 2, the optional H5 block, and the full T-sem block (the tier ordering across master seeds with re-drawn image pools) were not run; they remain pending. MAE with CLS pooling uses MAE's class token instead of the mean of its patch tokens; seeds 1 and 2 are two further random initializations of ResNet-50. In adults these variants entered only the sensitivity analyses of the primary test, and they are reported in the exploratory comparison on the same images (Table~\ref{tab:v2matched}). \\
Post hoc analyses after review & Specified on \VcPosthocDate, after two automated external reviews and after all results were known: by-image random slopes, exemplar distances, distance distributions, and H2 estimates with minimum detectable effects (Appendix~\ref{app:posthoc}). Outside every frozen plan; exploratory. \\
Exploratory comparison & Specified after the adult outcomes were known, run before the learner freeze, and not frozen separately (protocol and outputs share one commit). \\
H2 label for DINOv2-small & On ImageNet-HSJ, the DINOv2-small analysis (Section~\ref{sec:link}) reports the estimate as inconclusive (Holm-adjusted); the multi-encoder regions are unadjusted and classify the same estimate as support. \\
MAE positive control & Fitted with random intercepts only, because the model with by-person distance slopes did not pass the convergence and non-singularity check (the planned fallback). \\
Run interruptions & The human run exceeded its planned CPU budget without cuts; the agent run was suspended once and interrupted by a machine restart, and was resumed with the same frozen code. Results were unaffected. \\
\bottomrule
\end{tabular}
\caption{Deviations and Pending Analyses in the Multi-Encoder Analyses}
\label{tab:devtwo}
\end{table}

\subsection{History of Corrections to the Human-Data Analyses}
\label{app:history}

This appendix summarizes the history of the human-data analyses; the released deviation log documents each step with its scripts and outputs. The analysis plan specified random effects by participant identifier. Our first implementation grouped random effects by experiment $\times$ identifier, assuming that identifiers were reused across experiments for different people. An external review of the analyses pointed out that the original authors assigned identifiers after pooling the twelve experiments, so that one identifier is one person; we verified this in their anonymization script and re-estimated all H4 models, including the planned exploratory interaction, with random effects by person. The conclusions did not change. The person-level first-session sensitivity analysis was then redefined by each session's recorded start time. In the descriptive comparison of conditions, equivalent condition labels spelled differently across experiments were unified after an error was found; the H4 models were not affected. The test of the incremental contribution of distance within the taught domain was added after an internal critical review. The superordinate-only and trial-order models were specified after an automated external review of the manuscript, when the main results were already known; their specifications were written into the deviation log before they were run.

\section{Encoder Checkpoints}
\label{app:checkpoints}

\begin{table}[!htbp]
\centering\small
\setlength{\tabcolsep}{4pt}
\noindent\adjustbox{max width=\linewidth}{%
\begin{tabular}{llll}
\toprule
Encoder & Checkpoint & Revision & Embedding \\
\midrule
DINOv2-small (reference) & \texttt{facebook/dinov2-small} & \texttt{ed25f3a3} & CLS, last layer \\
DINOv2-base & \texttt{facebook/dinov2-base} & \texttt{f9e44c81} & CLS, last layer \\
CVCL & \texttt{wkvong/cvcl\_s\_dino\_resnext50\_embedding} & \texttt{f50eaa0c} & ResNeXt-50 vision encoder \\
SigLIP & \texttt{google/siglip-base-patch16-224} & \texttt{7fd15f06} & attention-pooled output \\
OpenCLIP & \texttt{laion/CLIP-ViT-B-32-laion2B-s34B-b79K} & \texttt{1a25a446} & projected image embedding \\
Supervised ViT & \texttt{google/vit-base-patch16-224} & \texttt{3f49326e} & CLS, last layer \\
Supervised ResNet-50 & \texttt{microsoft/resnet-50} & \texttt{34c2154c} & global average pool \\
MAE & \texttt{facebook/vit-mae-base} & \texttt{25b184be} & mean of patch tokens, mask ratio 0 \\
Random ResNet-50 (s0) & torchvision \texttt{resnet50(weights=None)}, seed 0 & -- & global average pool \\
Pixels & none & -- & resized, center-cropped pixels \\
\bottomrule
\end{tabular}}

\caption{Encoders and Pinned Checkpoint Revisions}
\label{tab:checkpoints}
\par\smallskip{\raggedright\footnotesize\textit{Note.} Revisions are the first eight characters of the Hugging Face commit pinned in the embedding script.\par}
\end{table}

Each representation used its own preprocessing, recorded in the released specification file. On square images, \VcCropList{} resize the shorter side and keep the central 224 of 256 pixels per side (87.5\%), whereas \VcFullList{} resize the whole image to $224 \times 224$; OpenCLIP resizes the shorter side to 224 and crops 224, which keeps the full frame of a square image. Pixels are the $32 \times 32$ RGB values after the DINOv2-style resize and crop. MAE uses the mean of its patch tokens (CLS pooling was a sensitivity variant), and the random ResNet-50 uses seed 0 (seeds 1 and 2 were sensitivity variants).

\clearpage
\section{Full Results by Representation}
\label{app:full}

\begin{table}[!htbp]
\centering\small
\setlength{\tabcolsep}{3.5pt}
\noindent\adjustbox{max width=\linewidth}{%
\begin{tabular}{llrlrlrlc}
\toprule
 & & \multicolumn{2}{c}{Positive control, 8 items} & \multicolumn{2}{c}{H4 replication, 24 items} & \multicolumn{2}{c}{Order moderation (primary)} & Region \\ \cmidrule(lr){3-4}\cmidrule(lr){5-6}\cmidrule(lr){7-8}
Encoder & Role & \multicolumn{2}{c}{$\beta_{z(d)}$} & \multicolumn{2}{c}{$\beta_{z(d)}$} & \multicolumn{2}{c}{$\hat\Delta$} & $\delta = .05/.10/.20$ \\
 & & \multicolumn{1}{c}{Est.} & \multicolumn{1}{c}{95\% CI} & \multicolumn{1}{c}{Est.} & \multicolumn{1}{c}{95\% CI} & \multicolumn{1}{c}{Est.} & \multicolumn{1}{c}{95\% CI} & \\
\midrule
DINOv2-small & Reference & \ensuremath{-}3.16 & \mbox{[\ensuremath{-}3.30,~\ensuremath{-}3.02]} & \ensuremath{-}3.23 & \mbox{[\ensuremath{-}3.29,~\ensuremath{-}3.16]} & 0.020 & \mbox{[\ensuremath{-}0.205,~0.274]} & I/I/I \\
DINOv2-base & Trained & \ensuremath{-}2.85 & \mbox{[\ensuremath{-}2.98,~\ensuremath{-}2.72]} & \ensuremath{-}2.86 & \mbox{[\ensuremath{-}2.91,~\ensuremath{-}2.80]} & \ensuremath{-}0.063 & \mbox{[\ensuremath{-}0.370,~0.244]} & I/I/I \\
CVCL & Trained & \ensuremath{-}0.62 & \mbox{[\ensuremath{-}0.71,~\ensuremath{-}0.52]} & \ensuremath{-}3.61 & \mbox{[\ensuremath{-}3.69,~\ensuremath{-}3.54]} & 0.065 & \mbox{[\ensuremath{-}0.058,~0.187]} & I/I/E \\
SigLIP & Trained & \ensuremath{-}3.89 & \mbox{[\ensuremath{-}4.06,~\ensuremath{-}3.72]} & \ensuremath{-}5.48 & \mbox{[\ensuremath{-}5.58,~\ensuremath{-}5.38]} & \ensuremath{-}0.044 & \mbox{[\ensuremath{-}0.352,~0.270]} & I/I/I \\
OpenCLIP & Trained & \ensuremath{-}3.99 & \mbox{[\ensuremath{-}4.16,~\ensuremath{-}3.83]} & \ensuremath{-}4.80 & \mbox{[\ensuremath{-}4.90,~\ensuremath{-}4.71]} & \ensuremath{-}0.057 & \mbox{[\ensuremath{-}0.284,~0.209]} & I/I/I \\
Supervised ViT & Trained & \ensuremath{-}4.30 & \mbox{[\ensuremath{-}4.46,~\ensuremath{-}4.13]} & \ensuremath{-}4.46 & \mbox{[\ensuremath{-}4.55,~\ensuremath{-}4.36]} & \ensuremath{-}0.034 & \mbox{[\ensuremath{-}0.318,~0.265]} & I/I/I \\
Supervised ResNet-50 & Trained & \ensuremath{-}2.65 & \mbox{[\ensuremath{-}2.77,~\ensuremath{-}2.52]} & \ensuremath{-}4.14 & \mbox{[\ensuremath{-}4.23,~\ensuremath{-}4.05]} & \ensuremath{-}0.026 & \mbox{[\ensuremath{-}0.172,~0.127]} & I/I/E \\
MAE & Trained & \ensuremath{-}0.20 & \mbox{[\ensuremath{-}0.30,~\ensuremath{-}0.10]}$^{a}$ & \ensuremath{-}7.56 & \mbox{[\ensuremath{-}7.66,~\ensuremath{-}7.45]} & 0.463 & \mbox{[\ensuremath{-}0.227,~1.180]} & I/I/I \\
\midrule
Random ResNet-50 (s0) & Control & \ensuremath{-}1.10 & \mbox{[\ensuremath{-}1.22,~\ensuremath{-}0.99]} & \ensuremath{-}2.68 & \mbox{[\ensuremath{-}2.73,~\ensuremath{-}2.63]} & \ensuremath{-}0.046 & \mbox{[\ensuremath{-}0.243,~0.148]} & I/I/I \\
Pixels & Control & 0.12 & \mbox{[0.04,~0.19]} & \ensuremath{-}2.43 & \mbox{[\ensuremath{-}2.47,~\ensuremath{-}2.39]} & 0.012 & \mbox{[\ensuremath{-}0.066,~0.088]} & I/E/E \\
\bottomrule
\end{tabular}}

\caption{Adult Word Extension by Representation}
\label{tab:v2adults}
\par\smallskip{\raggedright\footnotesize\textit{Note.} \VtPersons{} adults. Positive control: \VtPcNobs{} observations on the eight within-domain test images; a gate in the frozen plan, not the primary test. H4 replication: \VtHfourNobs{} observations on all 24 test images; largely driven by the contrast between in-domain and out-of-domain images, as for DINOv2-small. Order moderation (primary, confirmatory): experiment-level slope difference over \VtNexp{} experiments. Regions: I = inconclusive; E = equivalent within margin $\delta$. $^{a}$Fitted with random intercepts only, because the model with by-person distance slopes did not pass the convergence and non-singularity check; omitting the by-person slope tends to pull the slope toward zero relative to the other rows. Slopes are per standard deviation of each representation's distance (raw standard deviations in Appendix Table~\ref{tab:posthoc}). Regions are defined in Appendix~\ref{app:rules}. The controls are not gates. MAE with CLS pooling and random ResNet-50 seeds 1 and 2 entered only the sensitivity analyses of the primary test and are not shown.\par}
\end{table}

\begin{table}[!htbp]
\centering\small
\setlength{\tabcolsep}{3.5pt}
\noindent\adjustbox{max width=\linewidth}{%
\begin{tabular}{llrlrlrlcrc}
\toprule
 & & \multicolumn{2}{c}{C1 native (\%)} & \multicolumn{2}{c}{C1 far (\%)} & \multicolumn{2}{c}{D: partial $r$, 100 pairs} & Region D & D-all: partial $r$ & Region D-all \\ \cmidrule(lr){3-4}\cmidrule(lr){5-6}\cmidrule(lr){7-8}
Encoder & Role & \multicolumn{1}{c}{Est.} & \multicolumn{1}{c}{95\% CI} & \multicolumn{1}{c}{Est.} & \multicolumn{1}{c}{95\% CI} & \multicolumn{1}{c}{Est.} & \multicolumn{1}{c}{95\% CI} & SESOI $.10/.20/.30$ & 4{,}950 pairs & SESOI $.10/.20/.30$ \\
\midrule
DINOv2-small & Descriptive & 94.3 & \mbox{[92.2,~96.1]} & 53.0 & \mbox{[51.0,~55.0]} & \ensuremath{-}.50 & \mbox{[\ensuremath{-}.63,~\ensuremath{-}.33]} & S/S/S & \ensuremath{-}.60 & S/S/S \\
DINOv2-base & Confirmatory & 96.3 & \mbox{[94.6,~97.7]} & 54.4 & \mbox{[52.5,~56.4]} & \ensuremath{-}.57 & \mbox{[\ensuremath{-}.69,~\ensuremath{-}.42]} & S/S/S & \ensuremath{-}.69 & S/S/S \\
CVCL & Descriptive & 67.5 & \mbox{[63.6,~71.0]} & 42.8 & \mbox{[41.3,~44.2]} & .015 & \mbox{[\ensuremath{-}.18,~.21]} & I/N/N & \ensuremath{-}.09 & S/S/S \\
SigLIP & Confirmatory & 95.8 & \mbox{[94.6,~96.7]} & 53.4 & \mbox{[50.7,~56.3]} & \ensuremath{-}.30 & \mbox{[\ensuremath{-}.47,~\ensuremath{-}.11]} & S/S/S & \ensuremath{-}.48 & S/S/S \\
OpenCLIP & Confirmatory & 97.0 & \mbox{[96.2,~97.8]} & 52.4 & \mbox{[49.8,~55.2]} & \ensuremath{-}.35 & \mbox{[\ensuremath{-}.51,~\ensuremath{-}.16]} & S/S/S & \ensuremath{-}.52 & S/S/S \\
Supervised ViT & Confirmatory & 97.8 & \mbox{[97.0,~98.4]} & 51.8 & \mbox{[49.8,~54.0]} & \ensuremath{-}.65 & \mbox{[\ensuremath{-}.75,~\ensuremath{-}.52]} & S/S/S & \ensuremath{-}.68 & S/S/S \\
Supervised ResNet-50 & Confirmatory & 87.0 & \mbox{[84.3,~89.4]} & 49.5 & \mbox{[47.4,~51.6]} & \ensuremath{-}.30 & \mbox{[\ensuremath{-}.47,~\ensuremath{-}.11]} & S/S/S & \ensuremath{-}.39 & S/S/S \\
MAE & Confirmatory & 48.3 & \mbox{[45.2,~51.4]} & 38.8 & \mbox{[36.9,~40.6]} & \ensuremath{-}.004 & \mbox{[\ensuremath{-}.20,~.19]} & I/I/N & \ensuremath{-}.07 & S/S/S \\
\midrule
Random ResNet-50 (s0) & Control & 41.0 & \mbox{[38.4,~43.4]} & 38.4 & \mbox{[36.5,~40.3]} & .18 & \mbox{[\ensuremath{-}.02,~.37]} & N/N/N & .04 & N/N/N \\
Pixels & Control & 45.8 & \mbox{[42.7,~48.5]} & 38.7 & \mbox{[36.9,~40.5]} & \ensuremath{-}.14 & \mbox{[\ensuremath{-}.33,~.06]} & I/I/I & \ensuremath{-}.18 & S/S/S \\
\bottomrule
\end{tabular}}

\caption{Artificial Learners on CIFAR: Naming Accuracy and Distance (T, D) by Representation}
\label{tab:v2agentsD}
\par\smallskip{\raggedright\footnotesize\textit{Note.} C1: centroid naming accuracy for native and far CIFAR-10 tiers. D: partial correlation between inter-category distance and naming accuracy adjusting for WordNet distance, on the 100 pairs of the dissociation study; D-all: the same on all \VtDallN{} pairs (descriptive sensitivity; pairs share classes). Regions at SESOI .10/.20/.30: S = support, I = inconclusive, N = no support (Appendix~\ref{app:rules}); support requires a negative effect with unadjusted $p < .05$ and does not depend on the SESOI, so S/S/S does not mean that the effect exceeds .30. D-all regions treat the \VtDallN{} pairs as independent and are descriptive. DINOv2-small and CVCL: descriptive. Not run in the CIFAR analyses: MAE with CLS pooling and random ResNet-50 seeds 1 and 2.\par}
\end{table}

\begin{table}[!htbp]
\centering\small
\setlength{\tabcolsep}{3.5pt}
\noindent\adjustbox{max width=\linewidth}{%
\begin{tabular}{llrcrcrccccc}
\toprule
 & & \multicolumn{7}{c}{H1: Spearman $\rho$ and region} & \multicolumn{3}{c}{H2: region} \\ \cmidrule(lr){3-9}\cmidrule(lr){10-12}
 & & \multicolumn{2}{c}{THINGS} & \multicolumn{2}{c}{ImageNet-HSJ} & \multicolumn{2}{c}{CIFAR-10H} & CIFAR-10H & & & \\ \cmidrule(lr){3-4}\cmidrule(lr){5-6}\cmidrule(lr){7-8}
Encoder & Role & $\rho$ & Region & $\rho$ & Region & $\rho$ & Region & Mantel $p$ & THINGS & HSJ & CIFAR-10H \\
\midrule
DINOv2-small & Descriptive & .52 & S/S/S & .50 & S/S/S & .79 & S/S/S & $<$.001 & I/I/I & S/S/S & n/a$^{b}$ \\
DINOv2-base & Confirmatory & .52 & S/S/S & .55 & S/S/S & .68 & S/S/S & $<$.001 & I/I/I & I/I/I & n/a$^{b}$ \\
CVCL & Descriptive & .68 & S/S/S & .43 & S/S/S & .79 & S/S/S & $<$.001 & I/N/N & I/I/I & n/a$^{b}$ \\
SigLIP & Confirmatory & .64 & S/S/S & .50 & S/S/S & .89 & S/S/S & $<$.001 & I/I/I & S/S/S & n/a$^{b}$ \\
OpenCLIP & Confirmatory & .70 & S/S/S & .52 & S/S/S & .87 & S/S/S & $<$.001 & I/I/I & S/S/S & n/a$^{b}$ \\
Supervised ViT & Confirmatory & .39 & S/S/S & .56 & S/S/S & .66 & S/S/S & $<$.001 & I/I/I & I/I/I & n/a$^{b}$ \\
Supervised ResNet-50 & Confirmatory & .62 & S/S/S & .53 & S/S/S & .91 & S/S/S & $<$.001 & I/I/N & I/I/I & n/a$^{b}$ \\
MAE & Confirmatory & .29 & S/S/S & .23 & S/S/S & .44 & S/S/S & .015 & I/N/N & I/I/I & n/a$^{b}$ \\
\midrule
Random ResNet-50 (s0) & Control & .11 & S/S/S & .11 & S/S/S & .53 & S/S/S & .006 & I/I/I & S/S/S & n/a$^{b}$ \\
Pixels & Control & .15 & S/S/S & .20 & S/S/S & .29 & I/I/I & .0505 & I/N/N & I/I/I & n/a$^{b}$ \\
\bottomrule
\end{tabular}}

\caption{Artificial Learners and Human Similarity (H1, H2) by Representation}
\label{tab:v2agentsH}
\par\smallskip{\raggedright\footnotesize\textit{Note.} H1: Spearman correlation between encoder and human distances (THINGS, \VtHoneNThings{} pairs; ImageNet-HSJ, \VtHoneNHsj{}; CIFAR-10H, \VtHoneNCten{}), with region at SESOI .10/.20/.30 (support requires $\rho > 0$ and Mantel $p < .05$ and does not depend on the SESOI; Appendix~\ref{app:rules}). Mantel $p$ from 10{,}000 permutations. H2: unadjusted region for human distance predicting learner difficulty (the Holm-adjusted reference analysis of Section~\ref{sec:link} is inconclusive for DINOv2-small on ImageNet-HSJ); estimates in Appendix Table~\ref{tab:h2post}. $^{b}$Not applicable by design: H2 was defined only for THINGS and ImageNet-HSJ.\par}
\end{table}

\clearpage
\section{Decision Rules}
\label{app:rules}

This appendix restates the decision rules of the frozen plans; nothing in it was chosen after the data were seen.

\paragraph{Primary test of order moderation (adults).} For each representation $r$, item set (8 within-domain items or all 24), and experiment $x = 1, \dots, 12$, stage~1 fits the binomial mixed model
\[
\operatorname{logit} P(y = 1) = \beta_{0x} + \beta_x\, z_d + \textstyle\sum_c \gamma_{cx}\, C_c + \sum_c \eta_{cx}\, z_d C_c + \lambda_{x,\text{level}} + u_{\text{person}},
\]
where $z_d$ is the distance standardized within representation and item set over all experiments, $C_c$ are sum-to-zero contrasts for the four teaching conditions (so $\beta_x$ is the slope averaged over conditions), and the random-effects structure is the richest rung of a fixed ladder, from (1 + $z_d$~$\mid$~person) + (1~$\mid$~image) down to (1~$\mid$~person), that converged without singularity in all 12 experiments for all 10 representations. That rung was (1~$\mid$~person) for both item sets. Stage~2 computes
\[
\hat\Delta_r = \frac{1}{6}\sum_{x \in \mathcal{E}_{31}} \hat\beta_x - \frac{1}{6}\sum_{x \in \mathcal{E}_{13}} \hat\beta_x ,
\]
the mean slope in the six experiments with the one-example trial last minus the mean in the six with it first. The two-sided $p$ value is the share of the $\binom{12}{6} = 924$ assignments of the order labels for which $|\hat\Delta^{*}| \geq |\hat\Delta_r|$ (smallest attainable value $2/924 \approx .0022$). The 95\% and 90\% intervals collect the shifts $\delta_0$ (grid step .001) for which the test applied to $\hat\Delta_r - \delta_0$ (subtracting $\delta_0$ from the slopes of the one-example-last experiments) does not reject at $\alpha = .05$ and .10, respectively. Because order is confounded with blocking, a rejection would not identify order as the cause.

\paragraph{Regions for the primary test.} With 8 items and for each margin $\delta \in \{.05, .10, .20\}$ (log odds per standard deviation), the first region that applies is: (1) opposite sign, if the 95\% upper limit is below 0; (2) smaller than $\delta$, if the 90\% interval lies inside $(-\delta, \delta)$ (two one-sided permutation tests at $\pm\delta$); (3) holds, if the 95\% lower limit is above 0 and $\hat\Delta \geq \delta$; (4) attenuated, if the 95\% lower limit is above 0 and $\hat\Delta < \delta$; (5) inconclusive otherwise. A trained representation whose positive control failed was not interpretable. With 24 items there is no margin: opposite sign (upper limit $< 0$), positive (lower limit $> 0$), or inconclusive. The positive control required a negative Wald 95\% interval for the distance slope in the 8-item model of Section~\ref{sec:adults}. Claims that hold for all representations are intersection--union tests at each representation's own 95\% level; claims that the result depends on the representation are Holm-adjusted within two families per margin (trained representations other than DINOv2-small, two tests each; the two controls, two tests each). These $p$ values were never used to choose models.

\paragraph{Geometry calibration.} For each pair of representations, Pearson's $r$ was computed between their standardized distances over the 57{,}600 rows of the 8-item data, after residualizing both on level. Representations were clustered by average linkage (UPGMA) on $1 - r$ and the tree was cut at heights .3, .5, and .7. At a given cut the grouping \emph{discriminates} only if DINOv2-base falls in the group of DINOv2-small and neither control (random ResNet-50, pixels) shares a group with a trained encoder; otherwise that cut has ``no groups'' and no claim based on groups is made. DINOv2-base grouped with DINOv2-small at all three cuts, but a control shared a group with a trained encoder at every cut (MAE with pixels at .3; both controls with CVCL and MAE at .5 and .7), so calibration failed at every cut.

\paragraph{Regions for the CIFAR analyses (H1, H2, D).} With SESOI $\in \{.10, .20, .30\}$ on $|\rho|$ or the partial $|r|$, and without multiplicity adjustment, the first region that applies is: support, if $\rho > 0$ with Mantel $p < .05$ (H1), or the distance coefficient is negative with $p < .05$ (HC3 standard errors) and the semantic-distance coefficient has $p \geq .10$ or a partial $R^2$ below one third of the distance's (H2 and D); no support, if the 95\% interval lies entirely below the SESOI (H1), or excludes every value $\leq -\mathrm{SESOI}$ or shows an opposite effect with $p < .05$ (H2 and D); inconclusive otherwise. Support therefore does not depend on the SESOI, which affects only the boundary between no support and inconclusive. The plan anticipated that, with the half-width of the H2 interval near .35 on ImageNet-HSJ, no support would be nearly unattainable. Intervals for $r$ use the Fisher $z$ transformation with standard error $1/\sqrt{n - 4}$.

\paragraph{How $p < .05$ is used.} In the first plan, the six confirmatory tests (H1 on three datasets, H2 on two, and H4) were Holm-adjusted as one family; H5 and H6 were exploratory, with Benjamini--Hochberg correction. In the multi-encoder plans, per-representation regions use unadjusted $p < .05$ (or interval limits), and Holm adjustment applies only to the union claims described above. Observation-level likelihood-ratio tests, such as those of the trial-order models in Section~\ref{sec:adults}, are exploratory and do not enter any decision.

\clearpage
\section{Post Hoc Analyses after Review}
\label{app:posthoc}

The analyses in this appendix were specified on \VcPosthocDate, after two automated external reviews of an earlier draft and after all results were known. They are outside every frozen plan, are exploratory, and are reported as they came out. All use the frozen per-trial data and embeddings, with seed 42; a model counts as converged only under the criterion of the frozen plan (no error or convergence warning and a non-singular fit).

\paragraph{By-image random slopes.} The positive-control model of Section~\ref{sec:adults} (8 within-domain items) was refitted with a by-image random slope for distance, trying in order (1 + $z_d$~$\mid$~person) + (1 + $z_d$~$\mid$~image), the same with uncorrelated image effects, and the two corresponding structures with by-person intercepts only. The first structure converged for all ten representations (Table~\ref{tab:posthoc}). The average slope stayed negative for \VcImgTrSameSignWord{} of the eight trained encoders, but the image slopes varied widely (standard deviations from \VcImgSdMin{} to \VcImgSdMax{} log odds per standard deviation of distance) and the intervals widened accordingly: the interval excluded zero for \VcImgTrExclWord{} trained encoders and included it for \VcImgTrInclList. Under this model, the reference encoder would not have passed the positive control (\VcImgRzB{} [\VcImgRzLo,~\VcImgRzHi]). The random network kept an interval below zero, and the pixel slope turned negative with an interval that included zero. Models with an additional intercept for each image--condition cell did not converge for any representation. In the superordinate-only model of Section~\ref{sec:adults}, a by-image slope for distance removed the evidence that distance improved fit given condition ($\chi^2(1) = \VcSupImgChi$, $p = \VcSupImgP$, against \VcSupFrozenDistChi{} without it), whereas condition still improved fit given distance ($\chi^2(3) = \VcSupImgCondChi$). These results do not change the frozen analyses, but they show that the within-domain association between distance and extension is less certain at the level of images than the frozen models, which estimate a single slope for all images, suggest.

\paragraph{Exemplar distances.} For a test image $x$ and teaching images $s_1, \dots, s_k$, the nearest-exemplar distance is $\min_j d(x, s_j)$ and the mean-exemplar distance is $k^{-1}\sum_j d(x, s_j)$, where $d$ is the cosine distance between single embeddings; with one teaching image both equal the centroid distance. Each was standardized and entered in place of the centroid distance, with the random-effects structure that had converged for the centroid model, so that AIC values are comparable. Across the eight within-domain items, the mean-exemplar distance fitted better than the centroid distance for \VcDmeanBetterN{} of the ten representations and the nearest-exemplar distance for \VcDminBetterN; for DINOv2-small the differences were \VcPcDmeandAIC{} and \VcPcDmindAIC{} AIC points, with slopes of \VcPcDmeanB{} and \VcPcDminB{} (centroid: \VtPcRefB). In the superordinate-only model, teaching condition improved fit given each distance for every representation (smallest $\chi^2(3) = \VcSupCondChiMinAll$). Exemplar readouts therefore do not absorb the effect of the composition of the examples. Teaching sets differed sharply in dispersion (mean pairwise distance among the three teaching images for DINOv2-small: \VcDispSub{} after same-subtype and \VcDispSup{} after broad teaching), a property that the centroid distance and these exemplar readouts do not use; models that scale generalization by dispersion were not fitted.

\begin{table}[!htbp]
\centering
\caption{Post Hoc Adult Models (Exploratory)}
\label{tab:posthoc}
\small
\setlength{\tabcolsep}{3pt}
\noindent\adjustbox{max width=\linewidth}{%
\begin{tabular}{lcccccccc}
\toprule
 & \multicolumn{2}{c}{SD of $d$} & \multicolumn{2}{c}{Distance slope (8 items)} & & \multicolumn{2}{c}{$\Delta$AIC vs.\ centroid} & \\ \cmidrule(lr){2-3}\cmidrule(lr){4-5}\cmidrule(lr){7-8}
Representation & 8 & 24 & Frozen model (refit) & By-image slopes & SD$_{\text{image}}$ & Mean & Nearest & $\chi^2_{\text{cond}}$ \\
\midrule
DINOv2-small & .296 & .219 & \ensuremath{-}3.16 [\ensuremath{-}3.30, \ensuremath{-}3.02] & \ensuremath{-}2.79 [\ensuremath{-}5.88, 0.30] & 7.76 & \ensuremath{-}153 & +1562 & 18{,}198 \\
DINOv2-base & .331 & .230 & \ensuremath{-}2.85 [\ensuremath{-}2.98, \ensuremath{-}2.72] & \ensuremath{-}2.80 [\ensuremath{-}7.40, 1.80] & 12.88 & +78 & +1465 & 17{,}145 \\
CVCL & .117 & .159 & \ensuremath{-}0.62 [\ensuremath{-}0.71, \ensuremath{-}0.52] & \ensuremath{-}1.57 [\ensuremath{-}2.44, \ensuremath{-}0.71] & 2.14 & \ensuremath{-}581 & +180 & 18{,}862 \\
SigLIP & .094 & .105 & \ensuremath{-}3.89 [\ensuremath{-}4.06, \ensuremath{-}3.71] & \ensuremath{-}5.73 [\ensuremath{-}8.02, \ensuremath{-}3.44] & 5.77 & \ensuremath{-}32 & +1672 & 14{,}072 \\
OpenCLIP & .169 & .183 & \ensuremath{-}3.99 [\ensuremath{-}4.16, \ensuremath{-}3.83] & \ensuremath{-}3.48 [\ensuremath{-}5.64, \ensuremath{-}1.31] & 5.59 & \ensuremath{-}351 & +2226 & 11{,}645 \\
ViT (supervised) & .285 & .238 & \ensuremath{-}4.30 [\ensuremath{-}4.46, \ensuremath{-}4.13] & \ensuremath{-}4.87 [\ensuremath{-}6.44, \ensuremath{-}3.30] & 3.89 & +123 & +1333 & 7{,}019 \\
ResNet-50 (supervised) & .206 & .178 & \ensuremath{-}2.65 [\ensuremath{-}2.78, \ensuremath{-}2.52] & \ensuremath{-}3.71 [\ensuremath{-}4.37, \ensuremath{-}3.05] & 1.57 & +347 & +1808 & 5{,}042 \\
MAE & .011 & .012 & \ensuremath{-}0.20 [\ensuremath{-}0.31, \ensuremath{-}0.09]$^{a}$ & 0.15 [\ensuremath{-}3.61, 3.92] & 10.10 & +15 & \ensuremath{-}380 & 18{,}628 \\
Random ResNet-50 & .00028 & .00026 & \ensuremath{-}1.10 [\ensuremath{-}1.22, \ensuremath{-}0.98] & \ensuremath{-}1.85 [\ensuremath{-}3.29, \ensuremath{-}0.40] & 3.62 & \ensuremath{-}386 & +228 & 16{,}754 \\
Pixels & .120 & .115 & 0.12 [0.04, 0.20] & \ensuremath{-}0.13 [\ensuremath{-}1.24, 0.98] & 2.73 & \ensuremath{-}112 & \ensuremath{-}60 & 18{,}837 \\
\bottomrule
\end{tabular}
}
\par\smallskip{\raggedright\footnotesize\textit{Note.} SD of $d$: raw standard deviation of the cosine distance over the trial rows of the 8-item and 24-item data, the unit of each slope. Slopes: log odds per standard deviation of distance, with Wald 95\% CIs, in the 8-item positive-control model with (1 + $z_d$~$\mid$~person) + (1~$\mid$~image) (frozen) or (1 + $z_d$~$\mid$~person) + (1 + $z_d$~$\mid$~image) (by-image slopes). Intervals are from the post hoc refit and may differ from Table~\ref{tab:v2adults} in the second decimal. SD$_{\text{image}}$: standard deviation of the by-image slopes. $\Delta$AIC: AIC with the mean- or nearest-exemplar distance minus AIC with the centroid distance (negative favors the exemplar readout). $\chi^2_{\text{cond}}$: smallest likelihood-ratio statistic for condition given distance (3 df) in the superordinate-only model across the three readouts. $^{a}$Random intercepts only, as in Figure~\ref{fig:forest}.\par}
\end{table}

\begin{figure}[!htbp]
\centering
\includegraphics[width=\linewidth]{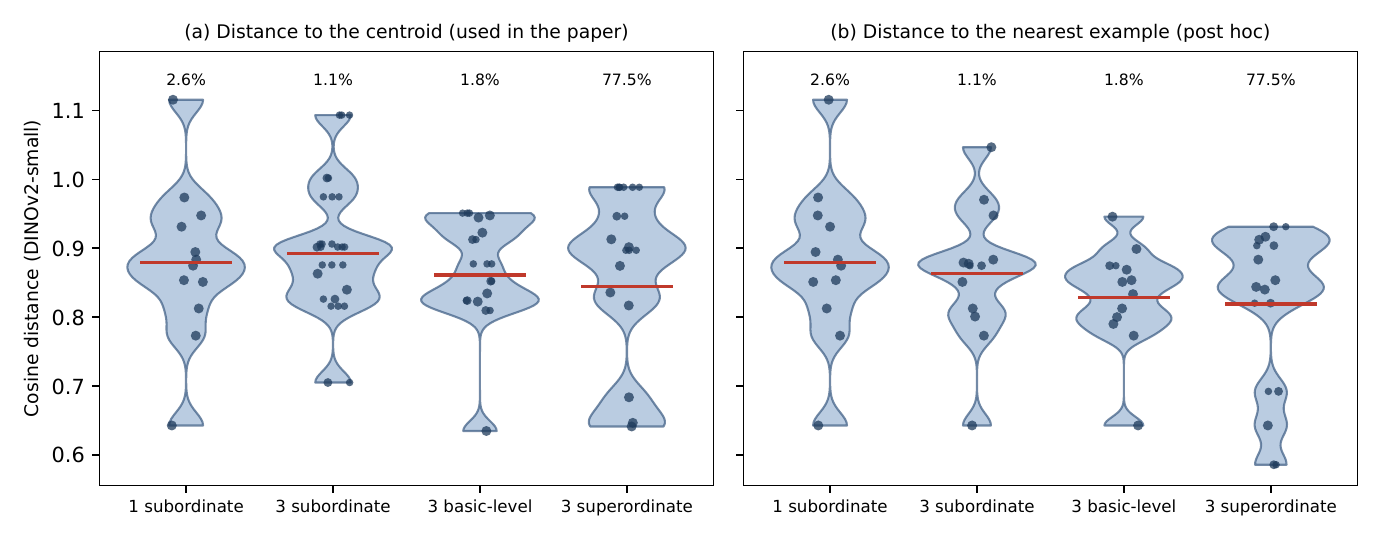}
\caption{Distances of Superordinate Test Items by Teaching Condition (DINOv2-small, Post Hoc). Distance of each superordinate test item to the centroid (left) and to the nearest teaching image (right), by teaching condition; points are the distinct image--teaching-set combinations, red lines the means, and the percentages above each violin the adults' selection rates. Mean centroid distance was \VcSupMeanSub{} (SD \VcSupSdSub) after same-subtype and \VcSupMeanSup{} (SD \VcSupSdSup) after broad teaching, while selection rose from \VcXSameSubPct\% to \VtXBroadPct\%.}
\label{fig:supdist}
\end{figure}

\paragraph{H2 estimates and detectable effects.} Table~\ref{tab:h2post} reports the H2 partial correlations for every representation with Fisher-$z$ intervals and HC3 $p$ values for the distance coefficient, the quantities behind the regions of Table~\ref{tab:v2agentsH}. With $n = 61$ and 32 pairs, the half-width of the 95\% interval at $r = 0$ is \VcMdeHalfThings{} and \VcMdeHalfHsj, and the smallest correlation detectable with 80\% power at two-sided $\alpha = .05$ is \VcMdeThings{} and \VcMdeHsj{}. The descriptive analysis on all mapped pairs (\VcHtwoAllThingsN{} for THINGS and \VcHtwoAllHsjN{} for ImageNet-HSJ) gave negative correlations for every representation, but its pairs share categories and are not independent, so it is not a test.

\begin{table}[!htbp]
\centering
\caption{Human Distance and Learner Difficulty (H2) by Representation (Post Hoc Summary)}
\label{tab:h2post}
\small
\setlength{\tabcolsep}{3pt}
\noindent\adjustbox{max width=\linewidth}{%
\begin{tabular}{lcccccc}
\toprule
 & \multicolumn{3}{c}{THINGS ($n = 61$)} & \multicolumn{3}{c}{ImageNet-HSJ ($n = 32$)} \\ \cmidrule(lr){2-4}\cmidrule(lr){5-7}
Representation & $r$ [95\% CI] & $p$ & all pairs $r$ & $r$ [95\% CI] & $p$ & all pairs $r$ \\
\midrule
DINOv2-small & \ensuremath{-}.10 [\ensuremath{-}.34, .16] & .374 & \ensuremath{-}.20 & \ensuremath{-}.41 [\ensuremath{-}.67, \ensuremath{-}.07] & .039 & \ensuremath{-}.18 \\
DINOv2-base & \ensuremath{-}.17 [\ensuremath{-}.41, .09] & .164 & \ensuremath{-}.22 & \ensuremath{-}.40 [\ensuremath{-}.66, \ensuremath{-}.05] & .052 & \ensuremath{-}.21 \\
CVCL & .08 [\ensuremath{-}.17, .33] & .561 & \ensuremath{-}.06 & \ensuremath{-}.39 [\ensuremath{-}.65, \ensuremath{-}.04] & .052 & \ensuremath{-}.05 \\
SigLIP & \ensuremath{-}.16 [\ensuremath{-}.40, .10] & .144 & \ensuremath{-}.27 & \ensuremath{-}.50 [\ensuremath{-}.73, \ensuremath{-}.18] & .011 & \ensuremath{-}.12 \\
OpenCLIP & \ensuremath{-}.20 [\ensuremath{-}.43, .06] & .088 & \ensuremath{-}.30 & \ensuremath{-}.45 [\ensuremath{-}.70, \ensuremath{-}.12] & .022 & \ensuremath{-}.13 \\
ViT (supervised) & \ensuremath{-}.08 [\ensuremath{-}.33, .17] & .476 & \ensuremath{-}.12 & \ensuremath{-}.39 [\ensuremath{-}.65, \ensuremath{-}.04] & .069 & \ensuremath{-}.24 \\
ResNet-50 (supervised) & \ensuremath{-}.04 [\ensuremath{-}.29, .22] & .749 & \ensuremath{-}.14 & \ensuremath{-}.38 [\ensuremath{-}.65, \ensuremath{-}.03] & .058 & \ensuremath{-}.10 \\
MAE & .13 [\ensuremath{-}.12, .38] & .229 & \ensuremath{-}.07 & \ensuremath{-}.27 [\ensuremath{-}.57, .10] & .131 & \ensuremath{-}.07 \\
Random ResNet-50 & \ensuremath{-}.06 [\ensuremath{-}.31, .20] & .655 & \ensuremath{-}.08 & \ensuremath{-}.35 [\ensuremath{-}.63, .00] & .032 & \ensuremath{-}.03 \\
Pixels & .08 [\ensuremath{-}.17, .33] & .487 & \ensuremath{-}.07 & \ensuremath{-}.28 [\ensuremath{-}.58, .08] & .143 & \ensuremath{-}.06 \\
\bottomrule
\end{tabular}
}
\par\smallskip{\raggedright\footnotesize\textit{Note.} $r$: partial correlation of human distance with naming accuracy, controlling semantic distance, on the covered pairs of the 100 dissociation pairs; 95\% CI from the Fisher $z$ transformation. $p$: HC3 $p$ value of the distance coefficient, used by the unadjusted regions. All pairs $r$: the same partial correlation over all mapped pairs, descriptive.\par}
\end{table}

\end{document}